\documentclass{article} 

\usepackage{amsmath,amsfonts,bm, amssymb}
\usepackage{algorithm}
\usepackage{algorithmic}

\def\eqref#1{equation~\ref{#1}}

\def\1{\bm{1}}

\DeclareMathAlphabet{\mathsfit}{\encodingdefault}{\sfdefault}{m}{sl}
\SetMathAlphabet{\mathsfit}{bold}{\encodingdefault}{\sfdefault}{bx}{n}

\usepackage{hyperref}
\usepackage{url}
\usepackage{graphicx}
\usepackage{subcaption}
\usepackage{booktabs}
\usepackage{multirow}
\usepackage{longtable}
\usepackage{comment}
\usepackage{pdflscape}
\usepackage{adjustbox}
\usepackage{rotating}
\usepackage{array}
\usepackage{ulem}
\usepackage{tabularx}
\usepackage{xcolor}
\usepackage{lmodern}
\usepackage{calc}
\usepackage{etoolbox}
\usepackage{iftex}
\usepackage[T1]{fontenc}
\usepackage[utf8]{inputenc}
\usepackage{textcomp} 

\usepackage{natbib}
\setcitestyle{authoryear,round,citesep={;},aysep={,},yysep={;}}
\usepackage{fancyhdr}

\title{Who Judges the Judge? LLM Jury-on-Demand: Building Trustworthy LLM Evaluation Systems}

\author{
\textbf{Xiaochuan Li}\thanks{These authors contributed equally to this work.},  
\textbf{Ke Wang}\footnotemark[1],
\textbf{Girija Gouda},\\
\textbf{Shubham Choudhary},
\textbf{Yaqun Wang},
\textbf{Linwei Hu},\\
\textbf{Joel Vaughan},
\textbf{Freddy Lecue} \\
\\
Wells Fargo Bank, N.A., USA \\
}

\begin{document}

\maketitle




\begin{abstract}
As Large Language Models (LLMs) become integrated into high-stakes domains, there is a growing need for evaluation methods that are both scalable for real-time deployment and reliable for critical decision-making. While human evaluation is reliable, it is slow and costly. Single LLM judges are biased, and static juries lack adaptability. To overcome these limitations, we propose LLM Jury-on-Demand - a dynamic, learning-based framework for scalable and context-aware evaluation. Our method trains a set of reliability predictors to assess when LLM judges will agree with human experts, leveraging token distributions, embeddings, and structural input features. This enables a fully adaptive evaluation where, for each data point, an optimal jury of the most reliable judges is dynamically selected and their scores are aggregated using their reliability as weights. Experiments on summarization and RAG benchmarks show that our dynamic jury system achieves significantly higher correlation with human judgment than both single-judge and static-jury baselines. These results highlight the promise of adaptive, learning-based juries for building scalable, more reliable and trustworthy evaluation systems for modern LLMs in high-stakes domains.\footnote{The views expressed in this paper are solely those of the authors and do not necessarily reflect the views of their affiliated institutions.}
\end{abstract}

\section{Introduction}

Large Language Models (LLMs) such as the GPT series, Llama, and Gemini
have demonstrated transformative capabilities, leading to their rapid
integration into critical, real-world applications
\citep{brown2020language, touvron2023llama,
team2023gemini}. As these models are deployed in high-stakes domains,
ensuring their outputs are reliable, safe, and aligned with human
expectations has become a paramount concern
\citep{shukla2025large, wang2023decodingtrust}. The gold
standard for assessing these qualities would be human evaluation, where
experts provide nuanced judgments. However, this process is notoriously
slow, expensive, and difficult to scale, making it impractical for the
rapid development cycles of modern AI
\citep{calderon2025alternative}. To overcome this
scalability bottleneck, the field historically relied on reference-based
automated metrics like BLEU and ROUGE, which measure lexical overlap
between the generated output and a ground-truth reference text
\citep{papineni2002bleu, lin2004rouge}. These methods
are now widely considered insufficient for capturing multifaceted
attributes like completeness, relevance, or groundedness in the
sophisticated outputs of modern generative models
\citep{zhang2019bertscore, cao2025toward}.

To address this evaluation gap, researchers have increasingly adopted the LLM-as-a-Judge paradigm, which leverages powerful language
models like GPT-4 to serve as scalable, automated evaluators
\citep{zheng2023judging, li2024llms, gu2024survey}.
While promising, this approach introduces a critical trade-off where the
scalability of a single LLM judge comes at the cost of reliability. The papers \citet{schroeder2024can},  \citet{li2024calibraeval} and \citet{baumann2025largelanguagemodelhacking} contain substantial evidence showing that single judges can be prone to systematic biases and inconsistencies, limiting their trustworthiness. A logical
evolution has been to employ a ``jury'' of multiple LLMs to improve
robustness \citep{feng2025one, verga2024replacing}.
However, these jury systems typically rely on static aggregation
methods, such as simple averaging. This fails to address a more
fundamental issue, as a judge\textquotesingle s expertise is
context-dependent and its reliability can change dramatically based on
the text being evaluated. This leaves a critical gap for a truly
adaptive evaluation system.

In this paper, we introduce LLM Jury-on-Demand, a novel framework that
bridges this gap by creating a dynamic, learning-based evaluation
system. Our work moves beyond static juries by training a system to
predict the reliability of each potential judge based on a rich set of
features extracted from the text. This allows our framework to perform a
fully adaptive evaluation where, for each data point, an optimal jury of
the most reliable judges is dynamically selected, and their scores are
aggregated using their reliability as weights. Our main contributions
are threefold:

\begin{itemize}
\item
  A new framework for adaptive LLM evaluation that
  demonstrates superior correlation with human judgment compared to
  single-judge and static-jury baselines.
  \item A method to predict LLM judge reliability at the instance level using text-based features.
  \item Extensive experiments and analyses across multiple tasks and datasets to validate the effectiveness of the proposed approach. 
\end{itemize}

\section{Related Work}\label{sec:related_work}

The evaluation of large language models is a rapidly evolving field, as captured in recent surveys mapping the transition from static benchmarks to more dynamic and automated evaluation frameworks \citep{cao2025toward}. Our work builds upon three key research areas: the LLM-as-a-Judge paradigm, the evolution from single judges to multi-model juries, and the broader concept of LLM performance prediction.

The LLM-as-a-Judge approach has become a scalable alternative to human annotation \citep{zheng2023judging}, with surveys documenting its widespread application and promising correlation with human preferences\citep{li2024llms, gu2024survey}. However, this paradigm has significant limitations. LLM judges exhibit biases, such as a preference for longer answers and sensitivity to the order in which responses are presented \citep{schroeder2024can}, and their judgments can be skewed by their own intrinsic style or pre-training data, which compromises the fairness and reliability of the evaluation \citep{li2024calibraeval}. These challenges motivate the need for more robust frameworks that can mitigate the inherent biases of a single judge.

To address these limitations, a growing body of work has explored using a ``jury'' of multiple LLMs, based on the insight that collaboration among diverse models can lead to more stable and reliable assessments \citep{feng2025one}. Initial work shows that simple ensembles, such as averaging the scores from a panel of smaller models, can outperform a single, larger model at a lower cost\citep{verga2024replacing, rahmani2024judgeblender}. More advanced methods have explored multi-agent frameworks where judges engage in peer-review or debate-like discussions to arrive at a consensus \citep{chu2024pre, zhao2024auto}. While a significant step forward, they typically rely on either static aggregation methods like simple voting or averaging or require complex and often unscalable conversational interactions. They do not account for the fact that a judge's expertise varies across different contexts, leaving a critical gap for systems that can adapt the jury's composition and weight to the specific context of the text being evaluated.

Our work is also grounded in LLM performance prediction. Studies have shown that it is possible to train a model to predict an LLM's performance on a given task by using features derived from the model and the task itself \citep{ye2023predictable}. Some approaches have even trained ``assessor'' models to predict when another model is likely to answer a question correctly, a concept that parallels our goal of predicting reliability \citep{schellaert2025analysing}. While these works validate the fundamental premise that LLM performance has learnable patterns, they typically focus on predicting general, task-level success rather than the instance-level reliability of an LLM acting as an evaluator. Our framework innovates by applying this concept to jury-based evaluation, enabling the dynamic selection and weighting of judges on a per-instance basis.

\section{Methodology}\label{sec:methodology}

In this section, we detail the architecture and components of LLM
Jury-on-Demand framework. Our framework is designed to produce more
reliable automated evaluations by shifting from a static to a dynamic,
learning-based process. The central hypothesis of our work is that an
LLM judge\textquotesingle s reliability is not fixed, but varies based
on the specific characteristics of the text it evaluates. Our system
models this variance by learning to predict when each judge is likely to
agree with human experts.

\subsection{Framework Overview}\label{sec:framework_overview}

\begin{figure}[h!]
\centering
\includegraphics[width=\textwidth]{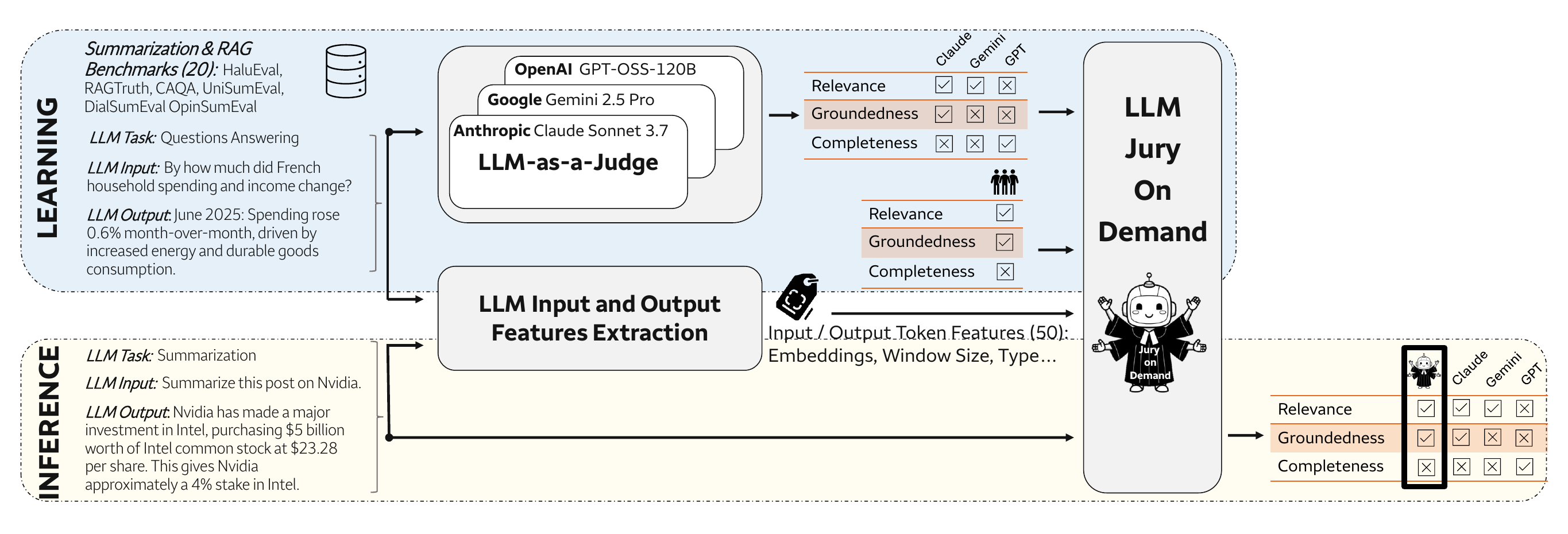}
\caption{Overview of the LLM Jury-on-Demand inference pipeline. The system extracts features from input texts to predict judge reliability, dynamically assembles a jury of the top $K$ most reliable judges for each instance, and calculates a final weighted score.}
\label{fig:pipeline}
\end{figure}

The LLM Jury-on-Demand framework operates through a multi-stage
pipeline, as shown in Fig. \ref{fig:pipeline}. The process begins with a set of input
texts related to the evaluation task. Crucially, the set of texts we
analyze depends on the task itself. For summarization tasks, the system
analyzes both the original source text and the model-generated output
summary. For Retrieval-Augmented Generation (RAG) tasks, it analyzes the
source text, the retrieved context, and the final generated answer. This
context-rich approach allows the system to capture a more complete
picture of the evaluation challenge.

These texts are first processed by a feature extraction module, which
extracts a wide range of textual and semantic signals. The resulting
feature vector for each instance serves as input to a suite of
pre-trained reliability prediction models. In the final stage, the
system leverages these reliability predictions to perform a fully
adaptive, per-instance evaluation. For each data point, a jury of a
pre-tuned size $K$ is dynamically assembled by selecting the judges
with the highest predicted reliability for that specific instance. The
final score is then computed as a weighted average of these selected
judges\textquotesingle{} raw scores, using their reliability predictions
as the weights.

\subsection{Feature Engineering for
Reliability
Prediction}\label{sec:FE_reliability_pred}

The foundation of our system is its ability to represent the evaluation
context through a rich set of predictive features. We hypothesize that
signals related to a text\textquotesingle s size, complexity, and
semantic content can reveal the scenarios in which different LLM judges
excel or struggle. The features are extracted from all available texts
(source, context, and output, as applicable to the task) and
concatenated into a single feature vector for each data point. Many of
these features are computed using Natural Language Toolkit (NLTK)
\citep{e3fe22bb6ca7471ea4b4acae1eb33f5d}. A complete
list of features is in the Appendix \ref{appendix:list-of-data-features}. Key feature
categories are:

\textbf{Text Size Features:} These include basic structural metrics such
as word count, sentence count, paragraph count, and the compression
ratio between the source and generated text.

\textbf{Special Words Features:} These count the occurrences of specific word types that can indicate the style or complexity of the text. Examples include counts of difficult words (words that have more than two syllables), named entities and  modality verbs (e.g. ``could'', ``should'') etc.

\textbf{Text Complexity Features:} These quantify readability and
ambiguity using established linguistic formulas. Examples include the
Flesch reading ease index
\citep{Kincaid1975DerivationON}, lexical diversity (the
variety of words used), and other measures of syntactic and semantic
ambiguity.

\textbf{Embedding-Related Features:} Embeddings encode text into a dense
vector representation \citep{Mikolov2013EfficientEO,
pennington-etal-2014-glove, devlin2019bert}. These capture the semantic
meaning and topic of the text. Top 10 PCA components
\citep{jolliffe2011principal} of text embeddings are
used as features. Additionally, we compute cosine similarity between
each text component's embedding and a set of predefined topic embeddings
(e.g. finance, technology), using these similarity scores as additional
topical relevance features.

\subsection{Learning to Predict Judge
Reliability}\label{sec:LP_judge_reliability}

Our framework learns judge reliability by training a dedicated machine
learning model for each specific evaluation context. That is, for each
potential judge, for each task (e.g., summarization), and for each
evaluation metric (e.g., completeness), we train a distinct model. The
purpose of this model is not to predict the evaluation score itself, but
rather to predict the probability that the corresponding judge will be
reliable on that metric for a given data point.

We frame this as a binary classification task to predict whether a
judge\textquotesingle s score will be ``good'' or ``bad''. To generate the
ground-truth labels for training, we compare each
judge\textquotesingle s score against a gold-standard human expert
score. First, we apply min-max normalization to all human and model
scores to scale them to a {[}0, 1{]} range. A judge\textquotesingle s
evaluation is then labeled as ``good'' (1) if its normalized score falls
within a predefined tolerance hyperparameter, $\tau$, of
the normalized human score. Otherwise, it is labeled as ``bad'' (0).

For each classification model, we use XGBoost, a gradient-boosted tree
algorithm known for its strong performance on tabular data
\citep{chen2016xgboost}. This approach is grounded in
the broader research area of LLM performance prediction, which has shown
that model performance can be learned from features
\citep{ye2023predictable, schellaert2025analysing}. At
inference time, our trained models output a probability score between 0
and 1, which we use as the predicted reliability.

\subsection{Assembling and Scoring the
Jury}\label{sec:assembling_jury}

The core of our framework is its ability to assemble an expert jury and
use its members\textquotesingle{} dynamically weighted opinions to
compute a final score. This process, which is applied for each
individual data point, involves two key mechanics: a reliability-based
jury selection algorithm and an instance-level dynamic weighting scheme.

\textbf{Jury Selection.} For each data point, we first use the suite of
pre-trained reliability models (described in Sec. \ref{sec:LP_judge_reliability}) to generate a
reliability score for each of the $N$ judges in our pool. A jury of a
pre-tuned size $K$ is then formed by simply selecting the $K$ judges
with the highest predicted reliability scores for that specific
instance. This approach allows the jury\textquotesingle s composition to
be completely dynamic, adapting to the unique characteristics of each
text.

\textbf{Dynamic Score Aggregation.} Once the instance-specific jury is
selected, the final evaluation score is calculated as a weighted average
of the raw scores from those $K$ jury members. The weights are derived
directly from the instance-specific reliability scores,
$[r_1,r_2 \dots,r_K]$.
Specifically, the weight for judge $i$ in the jury is calculated as
$w_i=r_i / (\sum_{j=1}^{K}r_j)$.
This dynamic, per-instance selection and weighting process allows our
system to prioritize the opinions of the most trustworthy judges for any
given text, which stands in contrast to prior systems that rely on
static juries and static aggregation methods like simple averaging
\citep{verga2024replacing}.

These two mechanics are the building blocks for our
system\textquotesingle s training and inference pipelines.

\subsection{System Training and
Inference}\label{sec:system-training-and-inference}

Our framework involves a one-time training and tuning phase to establish
an optimal configuration, which is then used in a repeatable inference
pipeline to evaluate new data points.

\textbf{Training and Hyperparameter Tuning.} The goal of the training
phase is to find the optimal hyperparameters that will generalize best.
This involves finding the single best jury size $K$ and the optimal
tolerance level $\tau$ for each individual judge's
reliability model. To do this, we first train a large pool of
reliability predictor models on our training data, covering all possible
combinations of judges and tolerance values. We then conduct a search
over the hyperparameter space of possible jury sizes ($K$) and
per-judge tolerance configurations. Each configuration is evaluated on a
held-out validation set. For every data point in the validation set, we
apply the \textit{Jury Selection} and \textit{Dynamic Score Aggregation}
mechanics described in Sec. \ref{sec:assembling_jury}. The configuration that yields the
highest Kendall\textquotesingle s Tau correlation with human scores
across the validation set is selected as the optimal configuration for
the final system.

\textbf{Inference Pipeline.} With the optimal configuration locked in
(i.e., a fixed $K$ and a set of optimal reliability models), the
system is ready to evaluate a new, unseen data point. For a new
instance, the system first uses the optimal reliability models to
predict an instance-specific reliability score for every potential judge
in the pool. It then applies the \textit{Jury Selection} and \textit{Dynamic
Score Aggregation} mechanisms to select the top $K$ judges and compute
the final, weighted score. 

\section{Experimental Setup}\label{sec:exp_setup}
To validate the effectiveness of our LLM
Jury-on-Demand framework, we conducted a series of experiments designed
to measure its performance against standard evaluation methods. This
section details the datasets, evaluation protocol, and implementation
specifics of our experiments.
\subsection{Datasets and Tasks}

Our evaluation spans two challenging natural language generation tasks: summarization and retrieval-augmented generation (RAG). For each task, we focus on evaluating three core metrics: groundedness, relevance, and completeness. These metrics are essential for assessing the quality and trustworthiness of generated text. Detailed definitions for each metric are provided in Appendix \ref{appendix:eval_metrics}.

To train our jury framework, we chose datasets with human annotations for these dimensions. We reviewed a diverse set of datasets for training, selecting 3-4 datasets per task by metric dimension to ensure coverage across various domains. To prevent any single dataset or any single metrics category from dominating a particular evaluation task, we applied stratified down-sampling where necessary. The details are provided in Appendix \ref{appendix:list-of-datasets}.

\subsection{Evaluation Protocol}\label{sec:eval_protocol}
Our experimental protocol is designed to ensure a fair and rigorous
comparison between our proposed system and relevant baselines.

\textbf{Evaluation Metric.} The primary metric for our experiments is
the Kendall\textquotesingle s Tau correlation coefficient \citep{10.1093/biomet/30.1-2.81}. This
non-parametric statistic measures the ordinal association between two
sets of rankings. In our context, it quantifies how well a
system\textquotesingle s evaluation scores align with the rankings
provided by human experts. A Kendall\textquotesingle s Tau value close
to 1 indicates strong agreement with human judgment.

\textbf{Judge Prompting Protocol.} To generate the raw scores for our
experiment, each potential judge model was prompted with a carefully
structured template. This template includes a system prompt to set the
judge\textquotesingle s persona (``You are a helpful, respectful and
honest assistant'') and a user prompt that defines the task, the specific
metric (e.g., Completeness), and the required scoring format. The
scoring scale was adapted to the task: for summarization, all judges
were instructed to provide a single integer score from 1 (lowest
quality) to 5 (highest quality); for RAG, a scale of 1 (lowest quality)
to 3 (highest quality) was used. The full prompt structure is provided
in the Appendix \ref{appendix:prompt-template}. Additionally, we conducted
an ablation study to analyse the effect on the system's resilience to slight
prompt variations as described in the Appendix \ref{appendix:ablation_prompt_effect}.

\textbf{Baselines.} To benchmark our system\textquotesingle s
performance, we established a judge pool consisting of 10 diverse LLMs.
This pool serves as the foundation for all evaluation methods compared
in our study and includes the following models: Claude 3.7 SONNET \citep{Claude3S}, Gemini 2.5 Pro \citep{comanici2025gemini}, Gemini 2.5 Flash, Gemini 2.0 Flash, GPT-OSS-20B \citep{agarwal2025gpt}, GPT-OSS-120B, Gemma 3-27B-IT \citep{team2025gemma}, Phi-4-reasoning \citep{abdin2025phi}, LLAMA-3.2-3B-Instruct \citep{grattafiori2024llama}, and DeepSeek-R1 \citep{guo2025deepseek}. From this pool, we formed two categories of
baselines:

\begin{enumerate}
\def\labelenumi{\arabic{enumi}.}
\item
  \textbf{Single-Judge Baselines:} The performance of each of the 10
  judges when used as a standalone evaluator.
\item
  \textbf{Static-Jury Baselines:} We compare against four distinct static jury formulations to rigorously test the benefits of dynamic selection:
\begin{itemize}
\item
  \textbf{Static Jury (Average-All):} The performance of a non-adaptive,
  naive jury that uses all 10 judges in the pool. For each data point,
  the final score for this baseline is the simple average of the raw
  scores from all 10 judges.
  \item  \textbf{Static Jury (Average-Top-K):} This baseline identifies the Top-$K$ best-performing single judges based on their Kendall's Tau on the validation set. The final score is the simple average of these Top-$K$ judges. The value of $K$ is tuned on the validation set.
  \item \textbf{Static Jury (Weighted-Regression):} A regression-based jury using all 10 judges. We train a linear regression model without intercept using human annotation scores as labels and single judge scores as features on the training set.
 \item \textbf{Static Jury (Weighted-Tau):} A performance-weighted average of all 10 judges. The weights are derived from each judge's validation Kendall's Tau, normalized using a softmax function.
\end{itemize}
\end{enumerate}

\subsection{Implementation Details}
This section provides the specific procedures used to configure and
train our system for the experiments.

First, we prepared the data for each task and metric by combining all
relevant source datasets. For example, to evaluate the completeness
metric for the summarization task, we aggregated the data from SummEval,
TLDR, and UniSumEval into a single, combined dataset. We then performed
a global 60-20-20 split on this combined data to create our final
training, validation, and holdout test sets, ensuring data from all
sources were represented in each split. As described in Sec. \ref{sec:LP_judge_reliability}, all
human and model scores were then normalized to a {[}0, 1{]} range to
ensure consistency.

Next, we determined the system's optimal configuration through a comprehensive hyperparameter tuning process on the combined validation set, as outlined in Sec. \ref{sec:system-training-and-inference}. We defined a search space for three key hyperparameter categories: the jury size $K$ (ranging from 2 to 9), the per-judge tolerance values $\tau$ used for training the reliability predictors, and the internal parameters of the XGBoost models. We exclude $K=1$ as it reduces the jury to a single judge selector, effectively duplicating the Single-Judge baseline paradigm. We also exclude $K=10$ (the full pool), as this configuration represents a static ensemble identical to the Average-All baseline, negating the benefit of dynamic selection. The sets of tolerance values were chosen to reflect the varying scales of the original human annotation scores across the different source datasets. For each candidate configuration, we evaluated its performance by applying the full per-instance jury selection and weighting pipeline (Sec. \ref{sec:assembling_jury}) to every data point in the combined validation set. The final optimal configuration was chosen based on which set of hyperparameters yielded the highest Kendall's Tau correlation on this combined validation set. This finalized configuration was then used for the final, unbiased evaluation on the locked-away test set.

Additionally, to assess robustness beyond validation-based tuning, we conduct ablation studies on jury size ($K$) and
tolerance ($\tau$), detailed in Appendix \ref{appendix:ablation_jury_size} and \ref{appendix:ablation_tolerance_levels}.
Jury size is varied from 1 to 9 across two representative tasks, Summarization-Completeness and RAG-Groundedness,
over 10 independent runs. As shown in Fig. \ref{fig:ablation_jury_size}, performance for both tasks follows a similar
pattern: accuracy improves as jury size increases, reaches an optimal range (around $K=5$–$8$), and then declines
slightly at very large sizes, indicating diminishing returns beyond the peak.

\section{Results and Analysis}

This section evaluates our LLM Jury-on-Demand framework against the single-judge and static-jury baselines. To ensure robustness, all experiments were repeated 10 times with different random seeds for data partitioning, and we report the mean and standard deviation of the results. Our analysis presents the main performance, both overall and at a granular dataset-level, followed by a diagnostic analysis of our system's internal mechanics, including feature importance and judge selection frequency.

\subsection{Performance Analysis}

\begin{figure}[h]
\centering
\includegraphics[width=1.0\textwidth]{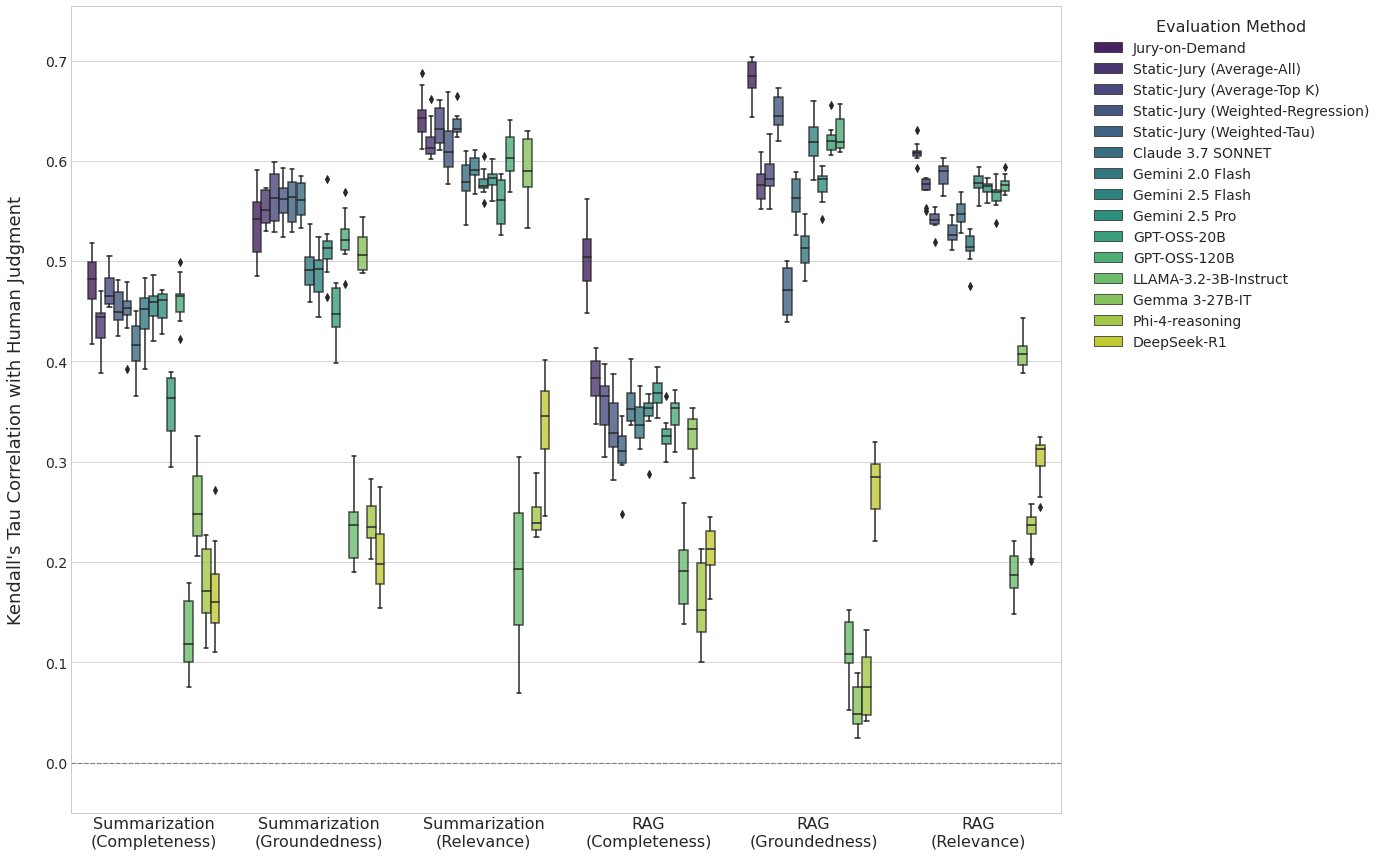}
\caption{Overall performance comparison over 10 runs. Boxplot of Kendall’s Tau correlation between each evaluation method’s scores and human judgements, aggregated across all datasets for the 6 task-metric combinations. Our Jury-on-Demand system achieves the highest median correlation in nearly all categories and shows the most robust performance.}
\label{fig:boxplot_kendalltau_overall}
\end{figure}

We assess performance by comparing the Kendall's Tau correlation of each method with human judgment on the test sets for each task-metric combination. The distribution of results from our 10 independent runs is summarized in Fig. \ref{fig:boxplot_kendalltau_overall}.

The results clearly demonstrate that our Jury-on-Demand framework consistently outperforms all baselines across every task and metric. In all six categories, our method achieves the highest mean correlation with human judgment, validating its effectiveness. For instance, in the challenging RAG-Groundedness task, our system achieves a mean Kendall's Tau of 0.68 (± 0.02). This represents a significant improvement not only over the simple Static Jury (Average-All) (0.58 ± 0.02) but also over the stronger optimized baselines, such as Static Jury (Weighted-Regression) (0.65 ± 0.02) and the strongest single judge for that task, GPT-OSS-120B (0.63 ± 0.02).

To provide a more detailed view of performance, we now analyze the results for the RAG-Completeness task at the individual dataset level. The granular results in Table \ref{tab:rag_completeness} confirm the findings from the overall analysis: our Jury-on-Demand system outperforms all four static jury baselines and the best-performing single judge on every individual dataset for this task. The performance lift is particularly pronounced on the ASQA dataset, indicating that our system is robust and its advantages are not an artifact of data aggregation but hold true at a more granular level. The full breakdown of results for all tasks and datasets is available in Appendix \ref{sec:full_res_1}. We also report statistical significance and effect sizes. Specifically, we perform one-sided Wilcoxon signed-rank tests \citep{wilcoxon1945individual} and compute Cliff’s delta \citep{articlecliff1993}, a non-parametric effect size metric that quantifies the difference between two groups. The results are presented in Appendix \ref{sec:full_res_1}.

Finally, the results also highlight the inherent unreliability of relying on a single LLM as an evaluator. As shown in Fig. \ref{fig:boxplot_kendalltau_overall}, the performance of single judges is highly variable. The ``best'' single judge changes from one task to another; for example, Claude 3.7 SONNET is often the strongest single judge for Summarization-Groundedness, but it is one of the weakest for Summarization-Completeness. This instability proves that there is no single ``best'' LLM judge, making a static choice of evaluator a risky and unreliable strategy.

The comparison between our Jury-on-Demand and the various static jury baselines isolates the benefit of our core contribution. While baselines like Static Jury (Weighted-Regression) are competitive, our dynamic system is consistently superior. This demonstrates that the primary performance gain comes not just from using a jury, but from the ability to dynamically select and weight its members based on the context of the input. The stability of this outperformance, demonstrated across 10 runs, underscores the reliability of our dynamic approach.

\begin{table}[h!]
\centering
\caption{Granular Performance on RAG-Completeness. Mean Kendall's Tau correlation ($\pm$ std. dev.) on the individual test sets across 10 runs. Our Jury-on-Demand system consistently outperforms all static baselines and the best single judge.}
\label{tab:rag_completeness}
\renewcommand{\arraystretch}{1.3}
\scriptsize
\begin{tabular}{|
>{\centering\arraybackslash}m{0.09\textwidth}|
>{\centering\arraybackslash}m{0.11\textwidth}|
>{\centering\arraybackslash}m{0.11\textwidth}|
>{\centering\arraybackslash}m{0.11\textwidth}|
>{\centering\arraybackslash}m{0.11\textwidth}|
>{\centering\arraybackslash}m{0.11\textwidth}|
>{\centering\arraybackslash}m{0.14\textwidth}|}
\hline
\textbf{Dataset} & \textbf{Jury-on-Demand} & \textbf{Static (Avg-All)} & \textbf{Static (Avg-TopK)} & \textbf{Static (W-Reg)} & \textbf{Static (W-Tau)} & \textbf{Best Single} \\
\hline
ALCE             & $0.47 \pm 0.07$         & $0.38 \pm 0.09$           & $0.28 \pm 0.08$            & $0.34 \pm 0.11$         & $0.23 \pm 0.10$         & $0.40 \pm 0.07$ (GPT-OSS-20B)  \\
ASQA             & $0.54 \pm 0.05$         & $0.38 \pm 0.05$           & $0.38 \pm 0.04$            & $0.36 \pm 0.04$         & $0.34 \pm 0.03$         & $0.42 \pm 0.03$ (Claude 3.7 SONNET) \\
QASPER           & $0.44 \pm 0.08$         & $0.41 \pm 0.08$           & $0.27 \pm 0.08$            & $0.35 \pm 0.07$         & $0.24 \pm 0.11$         & $0.43 \pm 0.07$ (GPT-OSS-120B) \\
\hline
\end{tabular}
\end{table}

\subsection{Analyzing the
Interaction Between Judges, Tasks, and Data
Attributes}\label{sec:diag_property}

\begin{figure}[htbp]
\centering
\begin{minipage}[b]{0.48\textwidth}
  \centering
  \includegraphics[width=\textwidth]{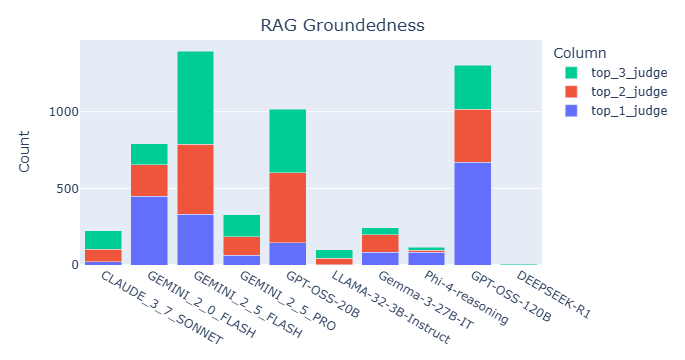}
\end{minipage}
\hfill
\begin{minipage}[b]{0.48\textwidth}
  \centering
  \includegraphics[width=\textwidth]{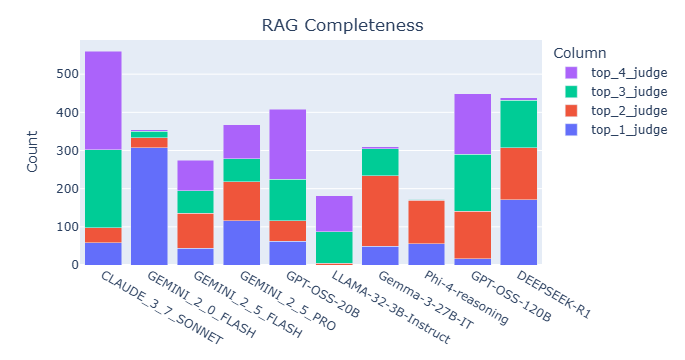}
\end{minipage}
\caption{Selection frequency of the judge in the jury. Top k
judge means that the judge has the k-th highest reliability score in the
jury. Claude 3.7 Sonnet and DeepSeek R1 are favored in
completeness, while Gemini 2.5 Flash is more often selected for groundedness.}
\label{fig:sel_freq}
\end{figure}

We begin by analyzing judge selection patterns within juries across different tasks and datasets. Fig. \ref{fig:sel_freq} summarizes the selection frequency of each judge for the RAG groundedness and RAG completeness tasks. The results reveal distinct preferences: Claude 3.7 Sonnet and DeepSeek R1 are frequently selected for completeness evaluation but are rarely chosen for groundedness. In contrast, Gemini 2.5 Flash is commonly selected for groundedness but appears less frequently in completeness evaluations. GPT OSS 20B and GPT OSS 120B are consistently selected across both metrics. A comprehensive comparison of judge selection across all tasks is in Appendix \ref{sec:full_res_2}.

We now examine how data properties impact judge performance. For illustration, we focus on two tasks: RAG groundedness and summarization completeness. The analysis for summarization completeness is in Appendix \ref{sec:full_res_2}. We select properties that rank among the most important features in the XGBoost model. For the summarization task, the key property is the compression ratio (i.e., the length of the summary divided by the length of the article). For the RAG task, the selected property is the character count of each response.

\begin{figure}[htbp]
\centering
\begin{minipage}[b]{0.48\textwidth}
  \centering
  \includegraphics[width=\textwidth]{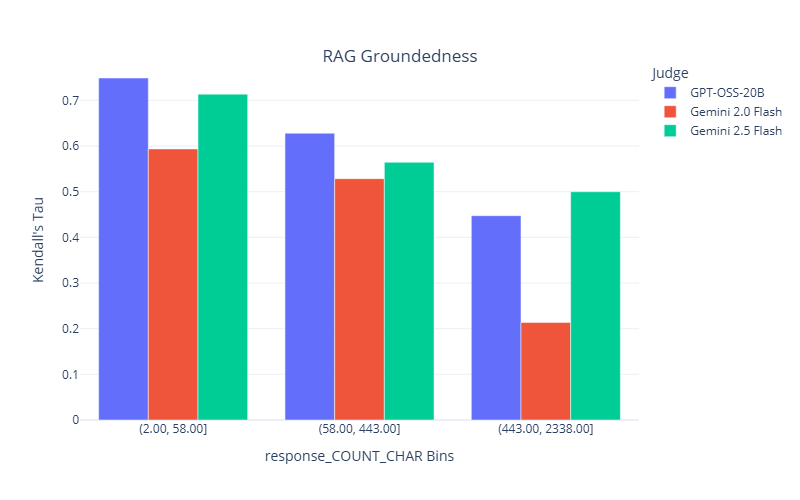}
\end{minipage}
\hfill
\begin{minipage}[b]{0.48\textwidth}
  \centering
  \includegraphics[width=\textwidth]{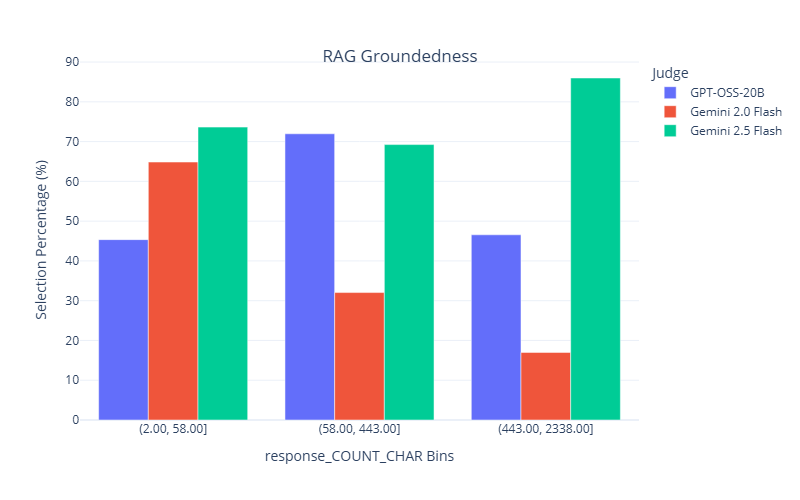}
\end{minipage}
\caption{RAG Groundedness analysis by response character count. (Left) Kendall's Tau correlation for three single judges across low, medium, and high response character count bins. (Right) The selection percentage of these judges in the final dynamic jury for data points within each bin. The analysis shows that judge performance degrades with longer responses, particularly for Gemini 2.0 Flash. Our system's jury selection adapts to this, heavily favoring the more reliable Gemini 2.5 Flash in the high-count bin.}
\label{fig:bin_rag}
\end{figure}

\begin{figure}[htbp]
\centering
\begin{minipage}[b]{0.48\textwidth}
  \centering
  \includegraphics[width=0.8\textwidth]{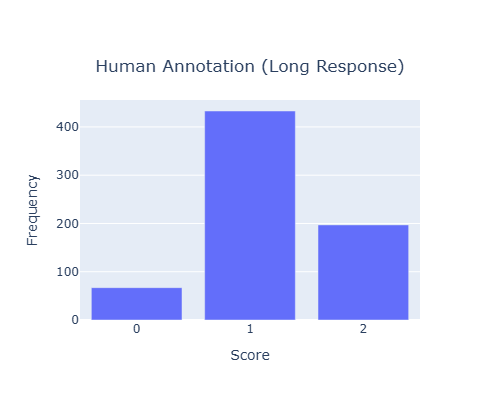}
\end{minipage}
\hfill
\begin{minipage}[b]{0.48\textwidth}
  \centering
  \includegraphics[width=0.8\textwidth]{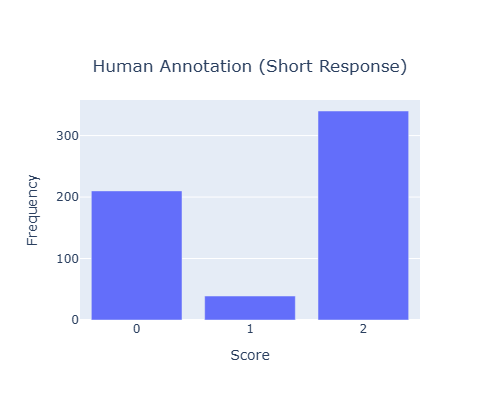}
\end{minipage}
\caption{RAG Groundedness: Distribution of annotation scores across response lengths. Longer responses tend to receive more score 1s (moderately ungrounded), while shorter responses are more often assigned scores of 0 (severely ungrounded) or 2 (fully grounded).}
\label{fig:score_summ}
\end{figure}

\begin{figure}[htbp]
\centering
\begin{minipage}[b]{0.48\textwidth}
  \centering
  \includegraphics[width=0.8\textwidth]{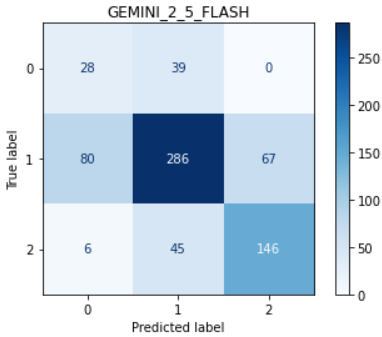}
\end{minipage}
\hfill
\begin{minipage}[b]{0.48\textwidth}
  \centering
  \includegraphics[width=0.8\textwidth]{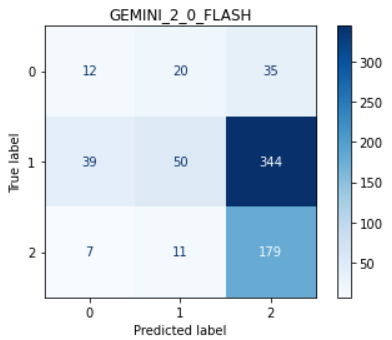}
\end{minipage}
\caption{RAG Groundedness: Confusion matrix for long responses. Gemini 2.0 Flash frequently assigns score 2 to ungrounded content, leading to reduced performance in this region.}
\label{fig:conf_long}
\end{figure}

\begin{figure}[htbp]
\centering
\begin{minipage}[b]{0.48\textwidth}
  \centering
  \includegraphics[width=0.8\textwidth]{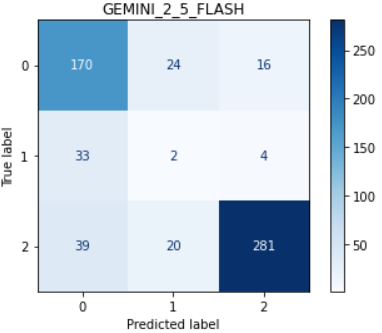}
\end{minipage}
\hfill
\begin{minipage}[b]{0.48\textwidth}
  \centering
  \includegraphics[width=0.8\textwidth]{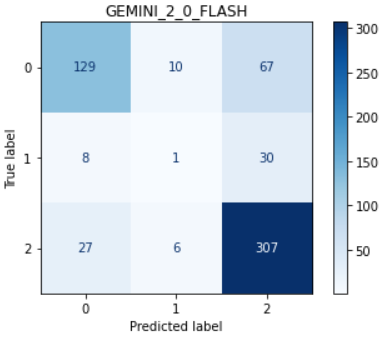}
\end{minipage}
\caption{RAG Groundedness: Confusion matrix for short responses. Gemini 2.0 Flash performs better in this region, as human annotation scores are predominantly 0 (severely ungrounded) or 2 (fully grounded), making them easier to distinguish.}
\label{fig:conf_short}
\end{figure}

To better illustrate the findings, we focus on three judges for each task. We begin with the RAG groundedness task. Fig. \ref{fig:bin_rag} shows model performance across bins of low, medium, and high response character counts. Two main observations emerge:
\begin{enumerate}
\def\labelenumi{\arabic{enumi}.}
\item
  All judges perform worse as the character count increases.
\item
   Gemini 2.0 Flash performs comparably to the others when the character count is low, but its performance drops significantly at higher character counts, especially compared to Gemini 2.5 Flash and GPT OSS 20B.
\end{enumerate}
To explain these trends, we examine the distribution of annotation scores in the low and high character count regions (see Fig. \ref{fig:score_summ}). In the high character count region, many responses receive a score of 1 (moderately ungrounded), whereas in the low character count region, most scores are either 0 (severely ungrounded) or 2 (fully grounded). It is easier for judges to distinguish between scores 0 and 2, but more difficult to differentiate between 1 and 2. Weaker judges, such as Gemini 2.0 Flash, particularly struggle with identifying ungrounded content.
Fig. \ref{fig:conf_long} and Fig. \ref{fig:conf_short} present the score confusion matrices for Gemini 2.0 Flash and Gemini 2.5 Flash for long and short responses. Compared to Gemini 2.5 Flash, Gemini 2.0 Flash assigns a disproportionately high number of score 2s, which explains its poor performance in the high character count region and relatively good performance in the low character count region. We also reviewed specific examples to understand why Gemini 2.0 Flash assigns score 2 even when the content is ungrounded. This analysis is provided in the Appendix \ref{sec:full_res_3_example}.

Finally, the jury selection dynamically aligns with model performance, as shown in Fig. \ref{fig:bin_rag}. This is most evident in the high character count bin, where the performance of Gemini 2.0 Flash drops significantly. Correspondingly, its selection percentage in our dynamic jury plummets to its lowest point, demonstrating that the reliability predictors correctly identify and avoid this weaker judge when it is unreliable. Conversely, Gemini 2.5 Flash maintains the highest performance in this high-count bin, and our system selects it for the jury in the vast majority of cases. GPT-OSS-20B also shows strong alignment in the medium character count bin, where it achieves its highest performance and is also the most selected judge. 

The framework also demonstrates robustness beyond simply picking the single best-performing judge. For instance, in the low character count bin, GPT-OSS-20B has the highest kendall's Tau. While it is selected frequently, Gemini 2.5 Flash, which also performs exceptionally well, is selected more often. This illustrates that the system does not rely on a single judge; rather, it identifies a pool of highly reliable judges for a given context and assembles an optimal jury from that pool.

These findings reinforce the importance of constructing dynamic juries that adapt to specific data characteristics and demonstrate the potential of our framework to predict judge reliability based on interpretable data properties.

Table \ref{tab:judge_vs_ftrs} presents the top five most important features for the judge reliability XGBoost model, as determined by permutation feature importance \citep{fisher2019all}, with the summarization groundedness task as an example. For illustration, we focus on Gemini 2.5 Pro and GPT-OSS-120B, while the complete results for all judges are in Appendix \ref{sec:full_res_2}. The analysis reveals the variation in feature importance across judges. For instance, embedding-related features are more influential for GPT-OSS-120B, suggesting that different judges rely on distinct data properties when assessing reliability.
We further aggregate the top five features that frequently appear across tasks, with results provided in Appendix \ref{sec:full_res_2}. The analysis reveals clear task-specific trends: text size-related features, such as word count and compression ratio, along with token entropy, are more prominent in RAG tasks. In contrast, embedding-based features, including PCA components and embedding similarity, play a more significant role in summarization tasks. These findings align with the ablation analysis in Appendix \ref{appendix:ablation}, which shows that removing embedding features leads to a greater performance drop in the summarization task compared to RAG. These observations imply that evaluation reliability is task-dependent and further demonstrate that our approach effectively links data characteristics to judge reliability, enabling more informed and adaptive jury construction across diverse evaluation scenarios.

\begin{table}[h!]
\caption{Top 5 important features for the summarization groundedness task. Gemini 2.5 Pro relies more on embedding-based features.}
\label{tab:judge_vs_ftrs}
\centering
\renewcommand{\arraystretch}{1.5}
\resizebox{\textwidth}{!}{
\scriptsize
\begin{tabular}{|p{0.1\textwidth}|p{0.12\textwidth}|p{0.12\textwidth}|p{0.17\textwidth}|p{0.15\textwidth}|p{0.15\textwidth}|}
\hline
\textbf{Judge} & \textbf{Feature 1} & \textbf{Feature 2} & \textbf{Feature 3} & \textbf{Feature 4} & \textbf{Feature 5} \\
\hline
Gemini 2.5 Pro & output compression & input pca7 & input reading index &  output embedding science & output pca10 \\ \hline
GPT-OSS-120B & output pca1 &  input pca1 & output compression & output embedding business & output embedding legal \\
\hline
\end{tabular}}
\end{table}

\subsection{Analysis of Jury Failure in Evaluation Tasks}

In this section, we investigate the conditions under which the dynamic jury fails to accurately assess model outputs. Given that jury performance is inherently dependent on the individual scoring behaviors of its constituent judges, our analysis aims to identify common regions and conditions where judges exhibit unreliable evaluations.

\begin{figure}[htbp]
\centering
\begin{minipage}[b]{0.48\textwidth}
  \centering
  \includegraphics[width=\textwidth]{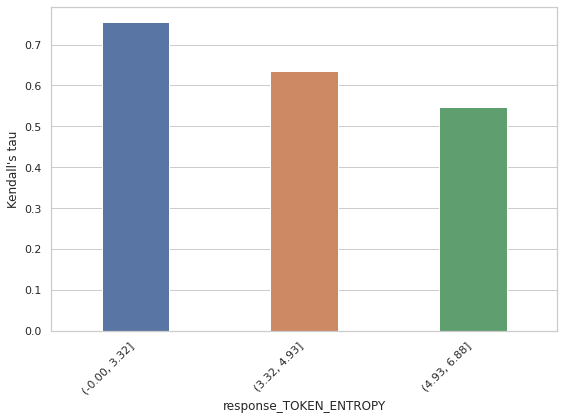}
\end{minipage}
\hfill
\begin{minipage}[b]{0.48\textwidth}
  \centering
  \includegraphics[width=\textwidth]{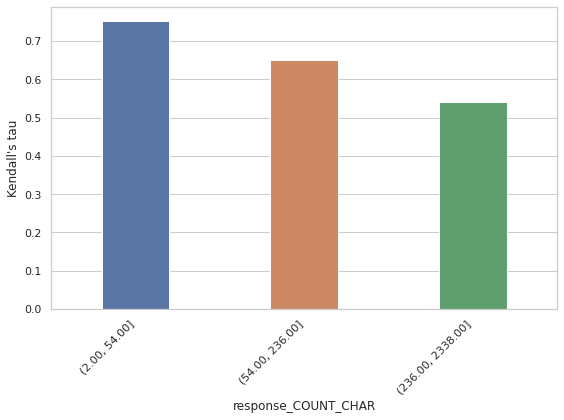}
\end{minipage}
\caption{Jury Performance Across Binned Segments of Token Entropy and Character Count in Generated Responses. Higher jury accuracy is observed when outputs are shorter or exhibit lower textual complexity.}
\label{fig:jury_bin}
\end{figure}

To uncover data attributes associated with jury failure, we train XGBoost models using a set of text attributes we measured as predictors and a binary response variable indicating jury success or failure. Feature importance is subsequently assessed via permutation-based methods to identify the most influential attributes affecting jury outcomes. For each of the top-ranked features, we conduct a binning analysis to evaluate jury performance across discrete intervals. Using the RAG groundedness task as a case study, we find that the most predictive features are related to the size and complexity of the generated text such as token entropy and character count. Figure \ref{fig:jury_bin} illustrates jury performance across bins of these two attributes, revealing a clear trend: as the length and entropy of generated text increase, the jury's ability to reliably assess groundedness diminishes. The complete results for top features identified in this task are presented in Appendix \ref{appendix:jury_fail_bin}.

\begin{figure}[htbp]
\centering
\begin{minipage}[b]{0.48\textwidth}
  \centering
  \includegraphics[width=\textwidth]{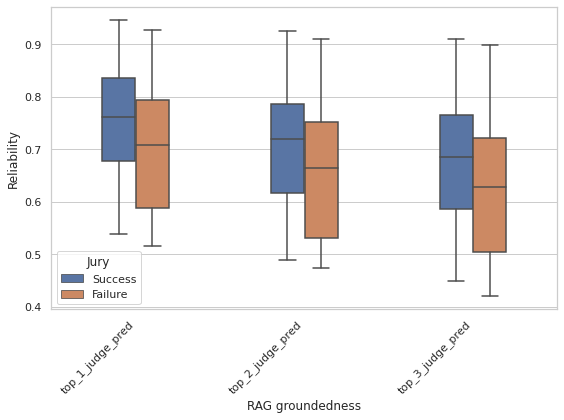}
\end{minipage}
\hfill
\begin{minipage}[b]{0.48\textwidth}
  \centering
  \includegraphics[width=\textwidth]{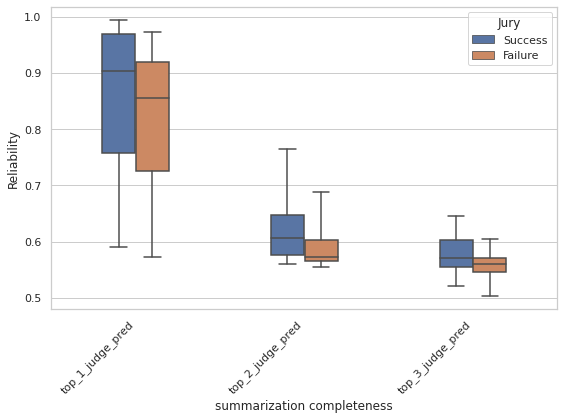}
\end{minipage}
\caption{Predicted Reliability Scores of Top Judges for Successful and Failed Jury Outcomes in RAG Groundedness and Summarization Completeness Tasks. Higher reliability scores are consistently observed in successful jury cases, validating the effectiveness of judge selection based on model-predicted performance.}
\label{fig:jury_relia}
\end{figure}

We further analyze the predicted reliability scores of the top-performing judges selected to form the jury, comparing cases of jury success versus failure. Using the RAG groundedness task as a representative example, we focus on the optimal jury configuration of three judges selected from a pool of ten for each evaluation instance.
Figure \ref{fig:jury_relia} presents a box plot summarizing the predicted reliability scores of the top three judges across both successful and failed jury outcomes. The results exhibit a consistent pattern: judges involved in successful jury decisions tend to have higher predicted reliability scores than those in failure cases. This trend is corroborated by similar findings in the summarization completeness task, detailed in Figure \ref{fig:jury_relia} as well. In that task, the optimal jury size is seven, we highlight the top three judges for clarity.

These observations validate the effectiveness of our jury construction strategy, which prioritizes the selection of judges based on their predicted reliability. Moreover, the results suggest that the XGBoost models used to estimate individual judge reliability accurately capture behavioral patterns. When the models predict high reliability, the corresponding judges typically perform well, contributing to successful jury outcomes. Conversely, when even the top-ranked judges struggle, the overall jury performance deteriorates.

\section{Conclusion}

In this work, we addressed the critical challenge of creating scalable
and reliable evaluation systems for Large Language Models. We introduced
LLM Jury-on-Demand, a novel framework that moves beyond the static
aggregation methods of prior jury-based systems by learning to predict
judge reliability and dynamically assembling an expert jury for each
data point. Our experimental results demonstrated that this adaptive
approach consistently outperforms both single-judge and static-jury
baselines in aligning with human expert judgement. This confirms our
central hypothesis that for automated evaluation to be trustworthy, it
must be context-aware and adaptive, rather than static.

While our results are promising, this work has several limitations that
open clear paths for future research. Our current framework relies on a
human-annotated dataset to train the reliability predictors; future work
could explore semi-supervised or self-supervised techniques to reduce
this dependency and enhance scalability. Furthermore, we conduct experiments to assess the framework’s ability to generalize beyond its training domains by training on a subset of domains and applying it to held-out domains. The results indicate partial generalization: certain learned patterns transfer effectively to new domains, while others do not (see Appendix \ref{appendix:unseen_domain} for details). These findings suggest that the framework’s generalizability is contingent on both the diversity of the training data and the characteristics of the unseen domains. As additional annotated data becomes available and incorporated into training, we anticipate that the framework’s capacity to generalize will improve, enabling broader applicability across diverse areas.

Another promising direction for future work is mitigating bias in judge scores. For example, certain judges may consistently assign higher or lower scores compared to human annotations. We explore score calibration in Appendix \ref{appendix:judge_calibration}. Our experiments show that calibration can sometimes improve alignment between judge scores and human annotations, but in other cases, it may amplify the bias. Addressing this challenge remains an open problem, and we plan to investigate more robust approaches in future work.

\bibliography{llm_checking}
\bibliographystyle{llm_checking}

\appendix

\section{List of Data
Features}\label{appendix:list-of-data-features}

\renewcommand{\arraystretch}{1.3}
\begin{longtable}{|p{4cm}|p{6cm}|p{2.5cm}|}\caption{Data features in judge reliability model.}\\
\hline
\textbf{Feature Name} & \textbf{Explanation} & \textbf{Category} \\
\hline
\endfirsthead

\hline
\textbf{Feature Name} & \textbf{Explanation} & \textbf{Category} \\
\hline
\endhead

COUNT\_WORD & Number of words in the context. & Text size \\\hline
COUNT\_CHAR & Number of characters in the context. & Text size \\\hline
COUNT\_SENTENCE & Number of sentences in the context. & Text size \\\hline
COUNT\_PARAGRAPH & Number of paragraphs in the context. & Text size \\\hline
CHAR\_COMPRESSION & The ratio of the number of characters in the output context (summary or answer) to those in the input context (article or cited context). & Text size \\\hline
WORD\_COMPRESSION & The ratio of the number of words in the output context (summary or answer) to those in the input context (article or cited context). & Text size \\\hline
NUM\_WORD\_ SENTENCE & Average number of words per sentence. & Text size \\\hline
NUM\_CHAR\_WORD & Average number of characters per word. & Text size \\\hline
DIFFICULT\_WORD & Number of difficult words in the context, defined as words with more than two syllables. & Special words \\\hline
STOP\_WORDS & Number of stop words in the context, such as “I”, “to”, “and”, “of”, etc. & Special words \\\hline
MODALITY & Number of modality verbs in the context, such as “can”, “could”, “should”, etc. & Special words \\\hline
NUMBER\_COUNT & Count of numbers in the context, such as date, time, percent etc. & Special words \\\hline
NAMED\_ENTITY & Count of named entities in the context, including person, organization, date, time, Geo-Political entity, location and money. & Special words \\\hline
FACTUAL\_DENSISTY & Number of entities divided by context length. & Special words \\\hline
NGRAM\_COUNT & Count of n(3)-grams in the context. & Special words \\\hline
NEGATION\_SENTENCE & Count of sentences with negation words such as “no”, “not”, “never”, etc. & Special words \\\hline
COUNT\_QUESTION & Number of questions in the context. & Special words \\\hline
TOKEN\_ENTROPY & Shannon entropy on token distribution. & Text complexity \\\hline
LEXICAL\_DIVERSITY & Number of unique words divided by the total number of words in the context. & Text complexity \\\hline
READING\_INDEX & Flesch reading ease index, measuring the difficulty of reading the context. & Text complexity \\\hline
NGRAM\_REPETITION & N(3)-gram repetition ratio in the context. & Text complexity \\\hline
SENTENCE\_ SIMILARITY & Average cosine similarity between each pair of sentence embeddings. & Text complexity \\\hline
SYNTACTIC\_ AMBIGUITY & The average number of syntactically ambiguous POS tags (IN, TO) across sentences. These tags indicate structural complexity or multiple possible parses. & Text complexity \\\hline
SEMANTIC\_ AMBIGUITY & The average number of WordNet senses per word across all sentences in the text. A higher average suggests more potential meanings and interpretive complexity. & Text complexity \\\hline
COREFERENCE\_CHAIN & The average number of pronouns per sentence. & Text complexity \\\hline
COREFERENCE\_ AMBIGUOUS & Number of pronoun-ambiguous sentences in the context. Sentences with more than one pronoun are classified as ambiguous. & Text complexity \\\hline
SYNTACTIC\_ANOMALY & Number of syntactic anomaly sentences in the context. A sentence is syntactic anomaly if either subject or verb is missing, or both are missing. & Text complexity \\\hline
RHETORICAL\_ STRUCTURE & Number of sentences with discourse markers (however, therefore) and rhetorical structure (moreover, in contrast, thus, instead, etc.). & Text complexity \\\hline
POLARITY & The polarity score of the context, measuring the emotional tone of the text. & Text complexity \\\hline
SUBJECTIVITY & The subjectivity score of the context, measuring the degree of personal opinion or factuality. & Text complexity \\\hline
PCA & Text embeddings are computed via mean pooling and reduced in dimensionality using PCA. The top 10 principal components are used as features. & Embedding \\\hline
Topic similarity & Cosine similarity between each text’s embedding and a set of predefined topic embeddings - market, bank, business, tech, education, politics, legal, sports, media, science. & Embedding \\
\hline
\end{longtable}

\section{Evaluation Metric Definitions}\label{appendix:eval_metrics}
For each task, we focus on evaluating three core metrics: groundedness, relevance, and completeness.

\textbf{Groundedness:} Assesses how well the output is supported by the context of the input. A grounded output accurately reflects the source information without introducing unsupported or fabricated content. This dimension is closely related to the concept of hallucination in language models.

\textbf{Relevance:} Measures the degree to which the output includes only essential and contextually appropriate information, avoiding extraneous or off-topic content. However, for RAG, annotated data which assesses output (answer) relevance is not readily available, so instead we check retrieval relevance. Specifically, how closely and thoroughly the retrieved context addresses the posed question. A context is considered relevant if it is clearly focused on the question and provides sufficient information to support a complete and accurate answer. Similarly, we can assess how relevant the context is with respect to the reference answer.

\textbf{Completeness:} Captures whether the output includes all critical information from the input context, ensuring comprehensive coverage.

\section{List of
Datasets}\label{appendix:list-of-datasets}

The datasets used for different evaluation metrics are listed in Table \ref{tab:listdataset_summary1} (summarization) and Table \ref{tab:listdataset_rag} (RAG). We prioritized datasets with annotated scores for completeness, groundedness, or relevance. However, annotated data for the completeness metric is relatively scarce. To address this, we simulate incomplete outputs by removing sentences from multi-sentence references and assigning scores accordingly. This approach is applied to the SummEval dataset for summarization and to all three datasets used in the RAG task.

\begin{small}
\renewcommand{\arraystretch}{1.3}
\begin{longtable}[]{@{}
  |>{\raggedright\arraybackslash}p{(\linewidth - 10\tabcolsep) * \real{0.1542}}|
  >{\raggedright\arraybackslash}p{(\linewidth - 10\tabcolsep) * \real{0.18}}|
  >{\raggedright\arraybackslash}p{(\linewidth - 10\tabcolsep) * \real{0.0674}}|
  >{\raggedright\arraybackslash}p{(\linewidth - 10\tabcolsep) * \real{0.26}}|
  >{\raggedright\arraybackslash}p{(\linewidth - 10\tabcolsep) * \real{0.0964}}|
  >{\raggedright\arraybackslash}p{(\linewidth - 10\tabcolsep) * \real{0.1724}}|@{}} \caption{List of Datasets for Summarization}
  \label{tab:listdataset_summary1}\\
 \toprule\noalign{}
\begin{minipage}[b]{\linewidth}\raggedright
Metric
\end{minipage} & \begin{minipage}[b]{\linewidth}\raggedright
Data
\end{minipage} & \begin{minipage}[b]{\linewidth}\raggedright
Size
\end{minipage} & \begin{minipage}[b]{\linewidth}\raggedright
Annotation (Ann.)
\end{minipage} & \begin{minipage}[b]{\linewidth}\raggedright
Ann. scale
\end{minipage} & \begin{minipage}[b]{\linewidth}\raggedright
Domain
\end{minipage} \\ 
\endhead
\bottomrule\noalign{}
\multirow{3}{=}{Completeness} & TL;DR
\citep{stiennon2020learning} & 1680 & Coverage score
measures how much important information from the original post is
covered & 1 to 7 & Reddit Discussion \\ \hline
& UniSumEval \citep{lee2024unisumeval} & 1623 &
Completeness ratio measures the proportion of key facts inferable from
the summary & 0 to 1 (fraction) & Nine different domains including
Wikihow, CNN/DM, GovReport, PubMed, etc. \\ \cline{2-6}
& SummEval \citep{fabbri2021summeval} & 793 & Original
data does not have completeness score. To create annotation for
completeness, we assign score 5 to the reference summary and remove
different proportions of sentences in the reference summary to create
incomplete summaries with scores 1 to 4 & 1 to 5 & CNN/DM \\
\hline
\multirow{4}{=}{Groundedness} & SummEval
\citep{fabbri2021summeval} & 1600 & Consistency score
measures the factual alignment between the summary and the summarized
source. A factually consistent summary contains only statements that are
entailed by the source document. & 1 to 5 & CNN/DM \\ \cline{2-6}
& TL;DR \citep{stiennon2020learning} & 1597 & Accuracy
score, measures to what degree the statements in the summary are stated
in the post & 1 to 7 & Reddit Discussion \\ \cline{2-6}
& DialSummEval \citep{gao-wan-2022-dialsummeval} & 1400
& Consistency score measures how well the summary aligns with the
dialogue in fact. It focuses on whether the summary contains factual
errors. & 1 to 5 & SAMSum (message dialogues) \\ \cline{2-6}
& UniSumEval \citep{lee2024unisumeval} & 1624 &
Faithfulness score measures the proportion of factually correct summary
sentences. & 0 to 1 (fraction) & 9 different domains including Wikihow,
CNN/DM, GovReport,PubMed, etc. \\
\hline
\multirow{4}{=}{Relevance} & OpinSummEval
\citep{shen2023opinsummeval} & 1400 & Aspect
relevance,~measures whether the mainly discussed aspects in the reviews
are covered exactly by the summary. It focuses on whether summary
correctly reflects the main aspects in the reviews. & 1 to 5 & Yelp
reviews \\ \cline{2-6}
& SummEval \citep{fabbri2021summeval} & 1600 &
Relevance score, measures selection of important content from the
source. The summary should include only important information from the
source document. & 1 to 5 & CNN/DM \\ \cline{2-6}
& DialSummEval \citep{gao-wan-2022-dialsummeval} & 1400
& Relevance score, measures how well the summary captures the key points
of the dialogue. It focuses on whether all and only the important
aspects are contained in the summary. & 1 to 5 & SAMSum (message
dialogues) \\ \cline{2-6}
& UniSumEval \citep{lee2024unisumeval} & 1623 &
Conciseness score, measures the proportion of summary sentences aligned
with the key-facts. & 0 to 1 (fraction) & Nine different domains
including Wikihow, CNN/DM, GovReport,PubMed, etc. \\
\hline
\end{longtable}
\end{small}

\newpage

\begin{small}
\renewcommand{\arraystretch}{1.3}
\begin{longtable}[]{@{}
  |>{\raggedright\arraybackslash}p{(\linewidth - 10\tabcolsep) * \real{0.1542}}|
  >{\raggedright\arraybackslash}p{(\linewidth - 10\tabcolsep) * \real{0.18}}|
  >{\raggedleft\arraybackslash}p{(\linewidth - 10\tabcolsep) * \real{0.0674}}|
  >{\raggedright\arraybackslash}p{(\linewidth - 10\tabcolsep) * \real{0.26}}|
  >{\raggedright\arraybackslash}p{(\linewidth - 10\tabcolsep) * \real{0.0964}}|
  >{\raggedright\arraybackslash}p{(\linewidth - 10\tabcolsep) * \real{0.1724}}|@{}}\caption{List of Datasets for RAG}
  \label{tab:listdataset_rag}\\
\toprule\noalign{}
\begin{minipage}[b]{\linewidth}\raggedright
Metric
\end{minipage} & \begin{minipage}[b]{\linewidth}\raggedright
Data
\end{minipage} & \begin{minipage}[b]{\linewidth}\raggedleft
Size
\end{minipage} & \begin{minipage}[b]{\linewidth}\raggedright
Annotation (Ann.)
\end{minipage} & \begin{minipage}[b]{\linewidth}\raggedright
Ann. scale
\end{minipage} & \begin{minipage}[b]{\linewidth}\raggedright
Domain
\end{minipage} \\
\endhead
\bottomrule\noalign{}
\multirow{3}{=}{Completeness} & ASQA
\citep{stelmakh-etal-2022-asqa} & 3231 & Original data
does not have completeness score. To create annotations for
completeness, we assign score 2 to the reference answer and remove
different proportions of sentences in the answer to create incomplete
answers with scores 0 and 1 & 0, 1, 2 & Wikipedia \\ \cline{2-6}
& ALCE \citep{gao2023enabling} & 593 & Original data
does not have completeness score. To create annotations for
completeness, we assign score 2 to the reference answer and remove
different proportions of sentences in the answer to create incomplete
answers with scores 0 and 2 & 0, 1, 2 & Wikipedia and Reddit \\  \cline{2-6}
& QASPER \citep{dasigi2021dataset} & 561 & Original data
does not have completeness score. To create annotations for
completeness, we assign score 2 to the reference answer and remove
different proportions of sentences in the answer to create incomplete
answers with scores 0 and 3 & 0, 1, 2 & NLP research papers \\  
\hline
\multirow{3}{=}{Groundedness} & RagTruth
\citep{niu-etal-2024-ragtruth} & 3206 & We assigned
scores 0 to 2 based on count of hallucination spans in the output & 0,
1, 2 & MS Marco \\ \cline{2-6}
& HaluEval \citep{li2023halueval} & 3000 & Whether
output contained hallucinated content~ & 0, 1 & HotPot-QA \\  \cline{2-6}
& CAQA \citep{hu-etal-2025-llms} & 3000 & Whether the
cited text supports the answer. There are 4 labels, supportive,
partially supportive, contradict and irrelevant & 0,1,2 & Knowledge
Graph generated questions \\ \hline
\multirow{3}{=}{Relevance} & MS MARCO
\citep{bajaj2016ms} & 3200 & Whether the cited text is
relevant to the question & 0, 1, 2, 3 & Bing queries \\  
\hline
& HotpotQA \citep{yang2018hotpotqa} & 3000 & Whether the
cited text is relevant to the question & 0,1,2 & Wikipedia-based QA \\  \cline{2-6}
& ALCE \citep{gao2023enabling} & 2896 & Whether the
cited text is relevant to the answer & 0,1,2 & Wikipedia and Reddit \\
\hline
\end{longtable}\end{small}

\section{Prompt
Template}\label{appendix:prompt-template}

We use the same prompt template across tasks. Below we list the prompt
for summarization (Table \ref{tab:prompt_summarize}) and RAG (Table \ref{tab:prompt_rag}), with dimension groundedness.

\begin{longtable}[]{@{}
  >{\raggedright\arraybackslash}p{(\linewidth - 0\tabcolsep) * \real{1.0000}}@{}}\caption{Prompt for summarization.}\label{tab:prompt_summarize}\\
\toprule\noalign{}
\begin{minipage}[b]{\linewidth}\raggedright
System prompt:\\
You are a helpful, respectful and honest assistant. Follow instructions
attentively. Do not add any extraneous information.\strut
\end{minipage} \\
\midrule\noalign{}
\endhead
\bottomrule\noalign{}
\endlastfoot
\begin{minipage}[t]{\linewidth}\raggedright
User prompt:

You will be given an input context followed by a corresponding summary.
Your task is to evaluate the quality of the summary with respect to
groundedness.\vspace{2mm}

Definition of Groundedness: Groundedness refers to how well the summary
is supported by the content of the input context. A grounded summary
should accurately reflect the information presented in the input context
without introducing unsupported claims.\vspace{2mm}

Provide a score for this summary on a scale from 1 (worst) to 5 (best).
Valid scores are: 1, 2, 3, 4, or 5.\vspace{2mm}

Output format:~\\
{[}score number{]}(on its own line, only one number here, no brackets or
letters or \textquotesingle score:\textquotesingle)\\
{[}explanation{]}(starting on the next line)\vspace{2mm}

Conversation: {source}\\
Summary: {output}\strut
\end{minipage} \\
\end{longtable}

\begin{longtable}[]{@{}
  >{\raggedright\arraybackslash}p{(\linewidth - 0\tabcolsep) * \real{1.0000}}@{}}\caption{Prompt for RAG.}\label{tab:prompt_rag}\\
\toprule\noalign{}
\begin{minipage}[b]{\linewidth}\raggedright
System prompt:\\
You are a helpful, respectful and honest assistant. Follow instructions
attentively. Do not add any extraneous information.\strut
\end{minipage} \\
\midrule\noalign{}
\endhead
\bottomrule\noalign{}
\endlastfoot
\begin{minipage}[t]{\linewidth}\raggedright
User prompt:

You will be given a question
(\textquotesingle Question\textquotesingle{} below) followed by a
response (\textquotesingle Response\textquotesingle{} below) for the
question. After that, cited background information is provided
(\textquotesingle Context\textquotesingle{} below). The response was
generated by a LLM based on the cited background information. Your task
is to evaluate the quality of the response with respect to groundedness.\vspace{2mm}

Definition of Groundedness: Groundedness refers to how well the response
is supported by the content of the cited background information. A
grounded response should accurately reflect the cited background
information without introducing unsupported claims.\vspace{2mm}

Provide a score for the response on a scale of 0 (bad), 1 (fair), or 2
(good). Valid scores are: 0, 1, or 2.\vspace{2mm}

Output format:~\\
{[}score number{]}(on its own line, only one number here, no brackets or
letters or \textquotesingle score:\textquotesingle)\\
{[}explanation{]}(starting on the next line)\vspace{2mm}

Question: {question}

Response: {response}

Context: {cited text}\strut
\end{minipage} \\
\end{longtable}

\section{Full Experimental
Results}\label{appendix:full-experimental-results}

We first show the jury performance, then analyze the interactions between judges, tasks and data properties.

\subsection{Full Results by Task and Dataset}\label{sec:full_res_1}

Tables \ref{tab:full_exp_res_summ_static} - \ref{tab:full_exp_res_rag_single} present the complete Kendall's Tau correlation results of our experiments. Due to the table size limit, we split the results for clarity. Tables \ref{tab:full_exp_res_summ_static} and \ref{tab:full_exp_res_rag_static} compare our Jury-on-Demand system against the four Static Jury baselines (Average-All, Average-TopK, Weighted-Regression, and Weighted-Tau). Tables \ref{tab:full_exp_res_summ_single} and \ref{tab:full_exp_res_rag_single} compare Jury-on-Demand against the 10 Single-Judge baselines. Each row corresponds to a specific evaluation set, either an ``Overall'' aggregation or an individual source dataset. All values represent the mean Kendall's Tau rank correlation coefficient ($\pm$ standard deviation) calculated across 10 independent runs. Higher values indicate better performance and stronger alignment with human judgment.

We also report statistical significance and effect sizes. Specifically, we perform one-sided Wilcoxon signed-rank tests \citep{wilcoxon1945individual} to compare the Tau differences between Jury-on-Demand and either static juries or single judges. The corresponding p-values are presented in Table \ref{tab:sigtest_summ_static} through Table \ref{tab:sigtest_rag_judge}. Among these p-values, 77$\%$ are statistically significant (p < 0.05). Values in parentheses represent Cliff’s delta \citep{articlecliff1993}, a non-parametric effect size metric that quantifies the difference between two groups, with Jury-on-Demand serving as the baseline. According to conventional thresholds, an effect size is considered large if it exceeds 0.47 and medium if it falls between 0.33 and 0.47. Across all Cliff’s delta values, 80$\%$ are classified as either large (70$\%$) or medium (10$\%$). This high proportion of significant p-values and substantial effect sizes indicates that, in most cases, Jury-on-Demand outperforms static baselines and single judges.

\begin{scriptsize}
\renewcommand{\arraystretch}{1.5}
\begin{longtable}{|
>{\raggedright\arraybackslash}m{1.8cm}
>{\raggedright\arraybackslash}m{1.4cm}
>{\raggedright\arraybackslash}m{2.0cm}
>{\raggedright\arraybackslash}m{2.0cm}
>{\raggedright\arraybackslash}m{2.0cm}
>{\raggedright\arraybackslash}m{2.0cm}|}
\caption{Summarization Results: Jury-on-Demand vs Static Jury baselines. Numbers in parentheses are standard deviation. \textbf{Bold} indicates the highest mean and its std in the row, \underline{underline} indicates the second highest mean and its std.}\label{tab:full_exp_res_summ_static}\\
\hline
\textbf{Data} & \textbf{Jury-on-Demand} & \textbf{Static Jury (Average-All)} & \textbf{Static Jury (Average-TopK)} & \textbf{Static Jury (Weighted-Regression)} & \textbf{Static Jury (Weighted-Tau)} \\
\hline
\endfirsthead

\hline
\textbf{Data} & \textbf{Jury-on-Demand} & \textbf{Static Jury (Average-All)} & \textbf{Static Jury (Average-TopK)} & \textbf{Static Jury (Weighted-Regression)} & \textbf{Static Jury (Weighted-Tau)}\\
\hline
\endhead

\multicolumn{6}{c}{\textbf{Completeness}} \\ \hline
Overall & \textbf{0.48}    & 0.44   & \underline{0.47}   & 0.45   & 0.45 \\ 
           & \textbf{(0.03)} & (0.02) & \underline{(0.02)} & (0.02) & (0.02) \\ \hline
SummEval & \textbf{0.72}   & 0.60    & 0.67   & \underline{0.69}    & 0.61 \\ 
                & \textbf{(0.05)} & (0.06) & (0.08) & \underline{(0.04)} & (0.05) \\ \hline
TL;DR    & 0.38    & 0.40   & \textbf{0.44}   & \underline{0.41}    & 0.41 \\ 
             & (0.08) & (0.05) & \textbf{(0.05)} & \underline{(0.02)} & (0.04) \\ \hline
UniSumEval & \textbf{0.66}   & 0.59    & \underline{0.63}    & 0.59   & 0.61 \\ 
                  & \textbf{(0.04)} & (0.03) & \underline{(0.04)} & (0.04) & (0.05) \\ \hline

\multicolumn{6}{c}{\textbf{Groundedness}} \\ \hline
Overall & 0.54    & 0.55   & \textbf{0.56}   & 0.56    & \underline{0.56} \\ 
           & (0.04) & (0.02) & \textbf{(0.03)} & (0.02) & \underline{(0.02)} \\ \hline
DialSummEval & 0.67   & 0.66    & \textbf{0.69}   & \underline{0.68}    & 0.68 \\ 
                     & (0.05) & (0.02) & \textbf{(0.02)} & \underline{(0.02)} & (0.02) \\ \hline
SummEval & 0.61    & \underline{0.65}  & 0.62     & 0.64   & \textbf{0.65} \\ 
                & (0.08) & \underline{(0.04)} & (0.05) & (0.03) & \textbf{(0.04)} \\ \hline
TL;DR    & 0.43    & 0.45   & 0.46    & \textbf{0.46}   & \underline{0.46} \\ 
             & (0.05) & (0.06) & (0.05) & \textbf{(0.05)} & \underline{(0.06)} \\ \hline
UniSumEval & 0.62    & 0.63   & 0.64    & \underline{0.64}   & \textbf{0.65} \\ 
                  & (0.07) & (0.07) & (0.07) & \underline{(0.07)} & \textbf{(0.06)} \\ \hline

\multicolumn{6}{c}{\textbf{Relevance}} \\ \hline
Overall & \textbf{0.64}    & 0.62   & 0.63    & 0.61   & \underline{0.64} \\ 
           & \textbf{(0.02)} & (0.02) & (0.02) & (0.03) & \underline{(0.01)} \\ \hline
DialSummEval & \textbf{0.69}   & 0.65    & 0.63   & 0.65    & \underline{0.66} \\ 
                     & \textbf{(0.04)} & (0.03) & (0.04) & (0.03) & \underline{(0.03)} \\ \hline
OpinSummEval & \textbf{0.46}   & 0.42   & 0.40    & 0.42    & \underline{0.44} \\ 
                      & \textbf{(0.06)} & (0.04) & (0.07) & (0.04) & \underline{(0.04)} \\ \hline
SummEval & \underline{0.71}    & 0.70   & \textbf{0.72}    & 0.69   & 0.70 \\ 
                & \underline{(0.07)} & (0.04) & \textbf{(0.05)} & (0.08) & (0.04) \\ \hline
UniSumEval & \textbf{0.42}   & 0.41    & 0.41    & 0.40   & \underline{0.42} \\ 
                  & \textbf{(0.09)} & (0.09) & (0.11) & (0.10) & \underline{(0.09)} \\ \hline
\end{longtable}
\end{scriptsize}

\begin{scriptsize}\renewcommand{\arraystretch}{1.5}
\begin{longtable}{|>{\raggedright\arraybackslash}p{0.8cm}
>{\raggedright\arraybackslash}m{0.7cm}
>{\raggedright\arraybackslash}m{0.7cm}
>{\raggedright\arraybackslash}m{0.7cm}
>{\raggedright\arraybackslash}m{0.7cm}
>{\raggedright\arraybackslash}m{0.7cm}
>{\raggedright\arraybackslash}m{0.7cm}
>{\raggedright\arraybackslash}m{0.7cm}
>{\raggedright\arraybackslash}m{0.7cm}
>{\raggedright\arraybackslash}m{0.7cm}
>{\raggedright\arraybackslash}m{0.7cm}
>{\raggedright\arraybackslash}m{0.7cm}|}
\caption{Summarization Results: Jury-on-Demand vs Single Judge baselines. Numbers in parentheses are standard deviation. Here Gemn. is Gemini and LL is LLAMA.
\textbf{Bold} indicates the highest mean and its std in the row, \underline{underline} indicates the second highest mean and its std.}
\label{tab:full_exp_res_summ_single}\\
\hline
\textbf{Data} & \textbf{Jury-on-Demand} & \textbf{Claude 3.7} & \textbf{Gemn. 2.0 Flash} & \textbf{Gemn. 2.5 Flash} & \textbf{Gemn. 2.5 Pro} & \textbf{GPT-OSS-20B} & \textbf{GPT-OSS-120B} & \textbf{LL-3.2} & \textbf{Gem\newline ma3} & \textbf{Phi4} & \textbf{Deep\newline Seek\newline R1} \\
 \hline
\endfirsthead

\hline
\textbf{Data} & \textbf{Jury-on-Demand} & \textbf{Claude 3.7} & \textbf{Gemn. 2.0 Flash} & \textbf{Gemn. 2.5 Flash} & \textbf{Gemn. 2.5 Pro} & \textbf{GPT-OSS-20B} & \textbf{GPT-OSS-120B} & \textbf{LL-3.2} & \textbf{Gem\newline ma3} & \textbf{Phi4} & \textbf{Deep\newline Seek\newline R1} \\
 \hline
\endhead

\multicolumn{12}{c}{\textbf{Completeness}} \\
\hline
Overall & \textbf{0.48}   & 0.42    & 0.45   & 0.46   & 0.45   & 0.35   & \underline{0.46}   & 0.12   & 0.26   & 0.18   & 0.17 \\
           & \textbf{(0.03)} & (0.03) & (0.03) & (0.02) & (0.02) & (0.04) & \underline{(0.02)} & (0.04) & (0.04) & (0.04) & (0.05) \\
\hline
Summ\newline Eval & \textbf{0.72}   & 0.57   & 0.63   & \underline{0.68}   & 0.58   & -0.10  & 0.67   & 0.12   & 0.33   & 0.27   & 0.68 \\
                             & \textbf{(0.05)} & (0.09) & (0.08) & \underline{(0.07)} & (0.11) & (0.15) & (0.08) & (0.14) & (0.16) & (0.06) & (0.04) \\
\hline
TL;DR    & 0.38   & 0.37   & 0.39   & \underline{0.40}    & \textbf{0.43}   & 0.34       & 0.40    & 0.09   & 0.14   & 0.01   & 0.05 \\
             & (0.08) & (0.07)  & (0.05) & \underline{(0.06)}  & \textbf{(0.06)} & (0.05) & (0.04) & (0.07) & (0.06) & (0.05) & (0.05) \\
\hline
Uni\newline Sum\newline Eval & \textbf{0.66}   & 0.52   & 0.54   & 0.59   & 0.52   & 0.62   & \underline{0.66}   & 0.29   & 0.55   & 0.29   & 0.41 \\
                                            & \textbf{(0.04)} & (0.04) & (0.07) & (0.04) & (0.05) & (0.04) & \underline{(0.04)} & (0.10) & (0.06) & (0.07) & (0.08) \\
\hline

\multicolumn{12}{c}{\textbf{Groundedness}} \\
\hline
Overall & \underline{0.54}   & \textbf{0.56}   & 0.49   & 0.49   & 0.51   & 0.45   & 0.52   & 0.23   & 0.51   & 0.24   & 0.20 \\
           & \underline{(0.04)} & \textbf{(0.02)} & (0.02) & (0.03) & (0.03) & (0.03) & (0.03) & (0.03) & (0.02) & (0.03) & (0.04) \\
\hline
Dial\newline Summ\newline Eval & 0.67   & 0.59   & 0.63   & \textbf{0.70}   & \underline{0.69}   & 0.65   & 0.66   & 0.33   & 0.64   & 0.24   & 0.33 \\
                                               & (0.05) & (0.04) & (0.03) & \textbf{(0.03)} & \underline{(0.03)} & (0.04) & (0.03) & (0.06) & (0.04) & (0.06) & (0.08) \\
\hline
Summ\newline Eval & 0.61    & \underline{0.63}    & 0.56   & 0.61   & \textbf{0.64}   & 0.56   & 0.60   & 0.21   & 0.59   & 0.17   & 0.12 \\
                             & (0.08) & \underline{(0.05)} & (0.06) & (0.04) & \textbf{(0.05)} & (0.06) & (0.04) & (0.09) & (0.06) & (0.09) & (0.10) \\
\hline
TL;DR    & \textbf{0.43}    & 0.39    & 0.41   & 0.40    & \underline{0.42}   & 0.29   & 0.39    & 0.10   & 0.42   & 0.11   & 0.13 \\
             & \textbf{(0.05)} & (0.04) & (0.05) & (0.05) & \underline{(0.04)} & (0.06) & (0.06) & (0.05) & (0.04) & (0.07) & (0.05) \\
\hline
Uni\newline Sum\newline Eval & \underline{0.62}   & 0.61   & 0.57   & 0.61   & \textbf{0.64}   & 0.61   & 0.59   & 0.10   & 0.55   & 0.31   & 0.24 \\
                                           & \underline{(0.07)} & (0.07) & (0.08) & (0.08) & \textbf{(0.07)} & (0.09) & (0.10) & (0.18) & (0.09) & (0.09) & (0.18) \\
\hline

\multicolumn{12}{c}{\textbf{Relevance}} \\
\hline
Overall & \textbf{0.64}    & 0.58   & 0.59   & 0.58   & 0.58   & 0.56   & \underline{0.61}   & 0.19   & 0.59   & 0.25   & 0.34 \\
           & \textbf{(0.02)} & (0.02) & (0.02) & (0.01) & (0.01) & (0.02) & \underline{(0.02)} & (0.08) & (0.03) & (0.02) & (0.05) \\
\hline
Dial\newline Summ\newline Eval & \textbf{0.69}   & 0.52   & 0.56   & 0.59   & 0.51   & 0.66   & \underline{0.68}   & 0.21   & 0.59   & 0.44   & 0.42 \\
                                               & \textbf{(0.04)} & (0.06) & (0.05) & (0.03) & (0.04) & (0.04) & \underline{(0.03)} & (0.09) & (0.03) & (0.04) & (0.08) \\
\hline
Opin\newline Summ\newline Eval & \textbf{0.46}   & 0.31   & 0.36   & 0.30    & 0.33   & 0.37    & 0.36   & 0.22    & \underline{0.39}   & 0.18   & 0.25 \\
                                                & \textbf{(0.06)} & (0.08) & (0.05) & (0.06) & (0.05) & (0.07) & (0.08) & (0.09) & \underline{(0.08)} & (0.09) & (0.07) \\
\hline
Summ\newline Eval & \textbf{0.71}   & 0.67    & 0.69    & 0.64    & 0.66   & 0.66   & 0.68   & 0.40    & \underline{0.69}   & 0.25   & 0.44 \\
                             & \textbf{(0.07)} & (0.05) & (0.06) & (0.09) & (0.05) & (0.05) & (0.09) & (0.17) & \underline{(0.05)} & (0.09) & (0.06) \\
\hline
Uni\newline Sum\newline Eval & \underline{0.42}    & \textbf{0.43}   & 0.37   & 0.38   & \textbf{0.43}   & 0.32   & 0.37   & 0.14   & 0.40   & 0.11   & 0.28 \\
                                            & \underline{(0.09)} & \textbf{(0.11)} & (0.09) & (0.10) & \textbf{(0.07)} & (0.07) & (0.09) & (0.12) & (0.10) & (0.08) & (0.12) \\
\hline

\end{longtable}
\end{scriptsize}


\begin{scriptsize}
\renewcommand{\arraystretch}{1.5}
\begin{longtable}{|
>{\raggedright\arraybackslash}m{1.8cm}
>{\raggedright\arraybackslash}m{1.4cm}
>{\raggedright\arraybackslash}m{2.0cm}
>{\raggedright\arraybackslash}m{2.0cm}
>{\raggedright\arraybackslash}m{2.0cm}
>{\raggedright\arraybackslash}m{2.0cm}|}
\caption{RAG Results: Jury-on-Demand vs Static Jury baselines. Numbers in parentheses are standard deviation. \textbf{Bold} indicates the highest mean and its std in the row, \underline{underline} indicates the second highest mean and its std.}\label{tab:full_exp_res_rag_static}\\
\hline
\textbf{Data} & \textbf{Jury-on-Demand} & \textbf{Static Jury (Average-All)} & \textbf{Static Jury (Average-TopK)} & \textbf{Static Jury (Weighted-Regression)} & \textbf{Static Jury (Weighted-Tau)} \\
\hline
\endfirsthead

\hline
\textbf{Data} & \textbf{Jury-on-Demand} & \textbf{Static Jury (Average-All)} & \textbf{Static Jury (Average-TopK)} & \textbf{Static Jury (Weighted-Regression)} & \textbf{Static Jury (Weighted-Tau)}\\
\hline
\endhead

\multicolumn{6}{c}{\textbf{Completeness}} \\ \hline
Overall & \textbf{0.50}    & \underline{0.38}   & 0.36   & 0.34    & 0.31 \\ 
           & \textbf{(0.03)} & \underline{(0.03)} & (0.03) & (0.03) & (0.03) \\ \hline
ALCE & \textbf{0.47}   & \underline{0.38}    & 0.28   & 0.34   & 0.23 \\ 
        & \textbf{(0.07)} & \underline{(0.09)} & (0.08) & (0.11) & (0.10) \\ \hline
ASQA & \textbf{0.54}   & \underline{0.38}    & 0.38   & 0.36   & 0.34 \\ 
         & \textbf{(0.05)} & \underline{(0.05)} & (0.04) & (0.04) & (0.03) \\ \hline
QASPER & \textbf{0.44}    & \underline{0.41}   & 0.27    & 0.35   & 0.24 \\ 
             & \textbf{(0.08)} & \underline{(0.08)} & (0.08) & (0.07) & (0.11) \\ \hline

\multicolumn{6}{c}{\textbf{Groundedness}} \\ \hline
Overall & \textbf{0.68}    & 0.58   & 0.59    & \underline{0.65}   & 0.47 \\ 
           & \textbf{(0.02)} & (0.02) & (0.02) & \underline{(0.02)} & (0.02) \\ \hline
CAQA & \textbf{0.68}   & 0.56    & 0.60   & \underline{0.62}    & 0.50 \\ 
         & \textbf{(0.03)} & (0.03) & (0.03) & \underline{(0.03)} & (0.05) \\ \hline
HaluEval & \textbf{0.77}   & 0.73    & \underline{0.74}   & 0.74    & 0.53 \\ 
             & \textbf{(0.02)} & (0.02) & \underline{(0.04)} & (0.04) & (0.05) \\ \hline
RagTruth & \textbf{0.57}   & 0.53    & 0.34    & \underline{0.55}   & 0.15 \\ 
              & \textbf{(0.03)} & (0.05) & (0.07) & \underline{(0.02)} & (0.08) \\ \hline

\multicolumn{6}{c}{\textbf{Relevance}} \\ \hline
Overall & \textbf{0.61}    & 0.57   & 0.54   & \underline{0.59}    & 0.53 \\ 
           & \textbf{(0.01)} & (0.01) & (0.01) & \underline{(0.01)} & (0.01) \\ \hline
ALCE & \textbf{0.61}   & 0.60    & 0.44   & \underline{0.60}   & 0.38 \\ 
        & \textbf{(0.03)} & (0.03) & (0.04) & \underline{(0.03)} & (0.03) \\ \hline
HotpotQA & \textbf{0.90}    & 0.86   & \underline{0.87}   & 0.85    & 0.86 \\ 
               & \textbf{(0.02)} & (0.02) & \underline{(0.03)} & (0.02) & (0.03) \\ \hline
MS MARCO & \textbf{0.46}    & 0.39   & \underline{0.45}   & 0.44    & 0.44 \\ 
                 & \textbf{(0.04)} & (0.04) & \underline{(0.04)} & (0.03) & (0.03) \\ \hline
\end{longtable}
\end{scriptsize}

\begin{scriptsize}\renewcommand{\arraystretch}{1.5}
\begin{longtable}{|>{\raggedright\arraybackslash}p{0.8cm}
>{\raggedright\arraybackslash}m{0.7cm}
>{\raggedright\arraybackslash}m{0.7cm}
>{\raggedright\arraybackslash}m{0.7cm}
>{\raggedright\arraybackslash}m{0.7cm}
>{\raggedright\arraybackslash}m{0.7cm}
>{\raggedright\arraybackslash}m{0.7cm}
>{\raggedright\arraybackslash}m{0.7cm}
>{\raggedright\arraybackslash}m{0.7cm}
>{\raggedright\arraybackslash}m{0.7cm}
>{\raggedright\arraybackslash}m{0.7cm}
>{\raggedright\arraybackslash}m{0.7cm}|}
\caption{RAG Results: Jury-on-Demand vs Single Judge baselines. Numbers in parentheses are standard deviation. Here Gemn. is Gemini and LL is LLAMA.
\textbf{Bold} indicates the highest mean and its std in the row, \underline{underline} indicates the second highest mean and its std.
}\label{tab:full_exp_res_rag_single}\\
\hline
\textbf{Data} & \textbf{Jury-on-Demand} & \textbf{Claude 3.7} & \textbf{Gemn. 2.0 Flash} & \textbf{Gemn. 2.5 Flash} & \textbf{Gemn. 2.5 Pro} & \textbf{GPT-OSS-20B} & \textbf{GPT-OSS-120B} & \textbf{LL-3.2} & \textbf{Gem\newline ma3} & \textbf{Phi4} & \textbf{Deep\newline Seek\newline R1} \\
 \hline
\endfirsthead

\hline
\textbf{Data} & \textbf{Jury-on-Demand} & \textbf{Claude 3.7} & \textbf{Gemn. 2.0 Flash} & \textbf{Gemn. 2.5 Flash} & \textbf{Gemn. 2.5 Pro} & \textbf{GPT-OSS-20B} & \textbf{GPT-OSS-120B} & \textbf{LL-3.2} & \textbf{Gem\newline ma3} & \textbf{Phi4} & \textbf{Deep\newline Seek\newline R1} \\
 \hline
\endhead

\multicolumn{12}{c}{\textbf{Completeness}} \\
\hline
Overall & \textbf{0.50}   & 0.36    & 0.34   & 0.35   & \underline{0.37}    & 0.33   & 0.35    & 0.19   & 0.33   & 0.16    & 0.21 \\
           & \textbf{(0.03)} & (0.02) & (0.02) & (0.02) & \underline{(0.02)} & (0.02) & (0.02) & (0.04) & (0.02) & (0.04) & (0.03) \\
\hline
ALCE    & \textbf{0.47}   & 0.38   & 0.34   & 0.30   & 0.34   & \underline{0.40}   & 0.36   & 0.16   & 0.33   & 0.08   & 0.32 \\
           & \textbf{(0.07)} & (0.09) & (0.10) & (0.10) & (0.08) & \underline{(0.07)} & (0.07) & (0.10) & (0.07) & (0.10) & (0.09) \\
\hline
ASQA    & \textbf{0.54}   & \underline{0.42}   & 0.38   & 0.33   & 0.34   & 0.33   & 0.38   & 0.22   & 0.36   & 0.13   & 0.20 \\
            & \textbf{(0.05)} & \underline{(0.03)} & (0.04) & (0.05) & (0.05) & (0.04) & (0.05) & (0.05) & (0.05) & (0.04) & (0.06) \\
\hline
QASPER  & \textbf{0.44}   & 0.31   & 0.23   & 0.41   & 0.35   & 0.40   & \underline{0.43}   & 0.05   & 0.32   & -0.02   & 0.15 \\
              & \textbf{(0.08)}& (0.07) & (0.08) & (0.08) & (0.08) & (0.08) & \underline{(0.07)} & (0.10) & (0.10) & (0.07)  & (0.14) \\
\hline

\multicolumn{12}{c}{\textbf{Groundedness}} \\
\hline
Overall & \textbf{0.68}   & 0.56   & 0.51   & 0.62   & 0.58   & 0.62   & \underline{0.63}   & 0.11   & 0.06   & 0.08   & 0.28 \\
           & \textbf{(0.02)} & (0.02) & (0.02) & (0.02) & (0.02) & (0.01) & \underline{(0.02)} & (0.03) & (0.02) & (0.03) & (0.03) \\
\hline
CAQA    & \textbf{0.68}  & 0.56   & 0.59    & 0.59   & 0.56   & 0.60   & \underline{0.60}   & 0.08   & 0.02   & 0.01   & 0.26 \\
            & \textbf{(0.03)}& (0.03) & (0.03) & (0.03) & (0.02) & (0.02) & \underline{(0.03)} & (0.06) & (0.02) & (0.03) & (0.04) \\
\hline
Halu\newline Eval & \underline{0.77}   & 0.67   & 0.52   & 0.74   & 0.73   & 0.76   & \textbf{0.78}   & 0.20   & 0.17   & 0.14   & 0.40 \\
                          & \underline{(0.02)} & (0.04) & (0.03) & (0.04) & (0.03) & (0.03) & \textbf{(0.03)} & (0.05) & (0.04) & (0.05) & (0.03) \\
\hline
Rag\newline Truth & \textbf{0.57}   & 0.40   & 0.30   & \underline{0.56}   & 0.49   & 0.51   & 0.52   & 0.12   & 0.14   & 0.07   & 0.14 \\
                           & \textbf{(0.03)} & (0.06) & (0.04) & \underline{(0.03)} & (0.05) & (0.05) & (0.05) & (0.05) & (0.05) & (0.09) & (0.07) \\
\hline

\multicolumn{12}{c}{\textbf{Relevance}} \\
\hline
Overall & \textbf{0.61}   & 0.55    & 0.51   & 0.58    & 0.57   & 0.57   & \underline{0.58}   & 0.19   & 0.41   & 0.23   & 0.30 \\
           & \textbf{(0.01)} & (0.01) & (0.02) & (0.01) & (0.01) & (0.01) & \underline{(0.01)} & (0.02) & (0.02) & (0.02) & (0.02) \\
\hline
ALCE    & \textbf{0.61}   & 0.51   & 0.53   & 0.57    & 0.58    & \underline{0.59}   & 0.58   & 0.11   & 0.43     & 0.13    & 0.28 \\
           & \textbf{(0.03)} & (0.03) & (0.04) & (0.04) & (0.05) & \underline{(0.03)} & (0.03) & (0.04) & (0.03) & (0.05) & (0.05) \\
\hline
Hotpot\newline QA & \textbf{0.90}    & 0.87   & 0.85    & 0.89   & 0.89   & 0.89   & \underline{0.90}   & 0.40   & 0.78   & 0.45   & 0.60 \\
                            & \textbf{(0.02)} & (0.03) & (0.02) & (0.02) & (0.02) & (0.02) & \underline{(0.01)} & (0.04) & (0.05) & (0.04) & (0.05) \\
\hline
MS\newline MARCO & \textbf{0.46}   & 0.40   & 0.40   & 0.43   & \underline{0.44}   & 0.39   & 0.43   & 0.14   & 0.17   & 0.10   & 0.19 \\
                             & \textbf{(0.04)} & (0.04) & (0.05) & (0.03) & \underline{(0.03)} & (0.04) & (0.03) & (0.06) & (0.05) & (0.05) & (0.06) \\
\hline
\end{longtable}
\end{scriptsize}


\begin{scriptsize}\renewcommand{\arraystretch}{1.5}
\begin{longtable}{|>{\raggedright\arraybackslash}p{1.8cm}>{\raggedright\arraybackslash}p{2.0cm}>{\raggedright\arraybackslash}p{2.0cm}>{\raggedright\arraybackslash}p{2.0cm}>{\raggedright\arraybackslash}p{2.0cm}|}\caption{Summarization: p-value of Wilcoxon test for the Tau difference between Jury-On-Demand and Static Jury baselines. Numbers in paranthesis () are effect size Cliff's delta.}\label{tab:sigtest_summ_static}\\
\hline
 \textbf{Data}  & \textbf{Static Jury (Average-All)} & \textbf{Static Jury (Average-TopK)} & \textbf{Static Jury (Weighted-Regression)} & \textbf{Static Jury (Weighted-Tau)} \\
 \hline
\endfirsthead

\hline
 \textbf{Data} & \textbf{Static Jury (Average-All)} & \textbf{Static Jury (Average-TopK)} & \textbf{Static Jury (Weighted-Regression)} & \textbf{Static Jury (Weighted-Tau)} \\
 \hline
\endhead

\multicolumn{5}{c}{\textbf{Completeness}} \\
\hline
Overall  & 0.001 & 0.161 & 0.024 & 0.019 \\
            & (0.76) & (0.26) & (0.56) & (0.64) \\\hline
Summ - Eval  & 0.001 & 0.116 & 0.116 & 0.001  \\
           & (0.88) & (0.34) & (0.44) & (0.86)  \\\hline
TL;DR     & 0.722 & 0.981 & 0.722 & 0.919  \\
            & (-0.08) & (-0.50) & (-0.06) & (-0.24) \\\hline
UniSum - Eval  & 0.002 & 0.014 & 0.001 & 0.005 \\
            & (0.82) & (0.46) & (0.86) & (0.54)  \\

\hline

\multicolumn{5}{c}{\textbf{Groundedness}} \\
\hline

Overall  & 0.958 & 0.981 & 0.995 & 0.997 \\
            & (-0.22) & (-0.42) & (-0.36) & (-0.38) \\\hline
Summ - Eval  & 0.958 & 0.652 & 0.968 & 0.976  \\
           & (-0.36) & (-0.14) & (-0.34) & (-0.42)  \\\hline
TL;DR     & 0.903 & 0.958 & 0.986 & 0.919  \\
            & (-0.14) & (-0.30) & (-0.36) & (-0.16) \\\hline
UniSum - Eval  & 0.500 & 0.813 & 0.688 & 0.903 \\
            & (-0.10) & (-0.10) & (-0.18) & (-0.32) \\\hline
Dial - Summ - Eval  & 0.313 & 0.958 & 0.784 & 0.688 \\
            & (0.08) & (-0.40) & (-0.22) & (-0.16) \\\hline

\multicolumn{5}{c}{\textbf{Relevance}} \\
\hline
Overall  & 0.002 & 0.080 & 0.002 & 0.116 \\
            & (0.60) & (0.14) & (0.64) & (0.24) \\\hline
Dial - Summ - Eval  & 0.001 & 0.002 & 0.007 & 0.032  \\
           & (0.62) & (0.74) & (0.54) & (0.46)  \\\hline
Opin - Summ - Eval     & 0.001 & 0.001 & 0.010 & 0.007  \\
            & (0.34) & (0.46) & (0.42) & (0.26) \\\hline
Summ - Eval     & 0.216 & 0.903 & 0.032 & 0.313  \\
            & (0.16) & (-0.08) & (0.18) & (0.14) \\\hline
UniSum - Eval  & 0.461 & 0.461 & 0.312 & 0.500 \\
            & (0.08) & (0.04) & (0.08) & (0.04)\\\hline

\end{longtable}
\end{scriptsize}


\begin{scriptsize}\renewcommand{\arraystretch}{1.5}
\begin{longtable}{|>{\raggedright\arraybackslash}p{0.8cm}>{\raggedright\arraybackslash}p{0.7cm}>{\raggedright\arraybackslash}p{0.7cm}>{\raggedright\arraybackslash}p{0.7cm}>{\raggedright\arraybackslash}p{0.7cm}>{\raggedright\arraybackslash}p{0.7cm}>{\raggedright\arraybackslash}p{0.7cm}>{\raggedright\arraybackslash}p{0.7cm}>{\raggedright\arraybackslash}p{0.7cm}>{\raggedright\arraybackslash}p{0.7cm}>{\raggedright\arraybackslash}p{0.7cm}|}\caption{Summarization: p-value of Wilcoxon test for the Tau difference between Jury-On-Demand and single judge. Here Gemn. is Gemini and LL is LLAMA. Numbers in paranthesis () are effect size Cliff's delta.}\label{tab:sigtest_summ_judge}\\
\hline
 \textbf{Data}  & \textbf{Claude 3.7} & \textbf{Gemn. 2.0 Flash} & \textbf{Gemn. 2.5 Flash} & \textbf{Gemn. 2.5 Pro} & \textbf{GPT-OSS-20B} & \textbf{GPT-OSS-120B} & \textbf{LL-3.2} & \textbf{Gem\newline ma3} & \textbf{Phi4} & \textbf{Deep\newline Seek\newline R1}\\
 \hline
\endfirsthead

\hline
 \textbf{Data}  \textbf{Claude 3.7} & \textbf{Gemn. 2.0 Flash} & \textbf{Gemn. 2.5 Flash} & \textbf{Gemn. 2.5 Pro} & \textbf{GPT-OSS-20B} & \textbf{GPT-OSS-120B} & \textbf{LL-3.2} & \textbf{Gem\newline ma3} & \textbf{Phi4} & \textbf{Deep\newline Seek\newline R1}\\
 \hline
\endhead

\multicolumn{11}{c}{\textbf{Completeness}} \\
\hline
 			Overall  & 0.002 & 0.024 & 0.032 & 0.024 & 0.001 & 0.042 & 0.001 & 0.001 & 0.001 & 0.001\\
           & (0.92) & (0.62) & (0.50) & (0.50) & (1.00) & (0.40) & (1.00) & (1.00) & (1.00) & (1.00)\\\hline
Summ - Eval  & 0.005 & 0.014 & 0.096 & 0.005 & 0.001 & 0.096 & 0.001 & 0.001 & 0.001  & 0.042\\
           & (0.92) & (0.66) & (0.32) & (0.76) & (1.00) & (0.46) & (1.00) & (1.00) & (1.00)  & (0.56)\\\hline
TL;DR     & 0.312 & 0.500 & 0.652 & 0.920 & 0.065 & 0.615 & 0.001 & 0.001 & 0.001 & 0.001\\
            & (0.04) & (0.02) & (-0.12) & (-0.34) & (0.44) & (0.04) & (1.00) & (0.98) & (1.00) & (1.00)\\\hline
UniSum - Eval  & 0.001 & 0.001 & 0.002 & 0.001 & 0.003 & 0.042 & 0.001 & 0.002 & 0.001 & 0.001\\
            & (1.00) & (0.90) & (0.78) & (1.00) & (0.62) & (0.12) & (1.00) & (0.94) & (1.00) & (1.00)\\

\hline

\multicolumn{11}{c}{\textbf{Groundedness}} \\
\hline

Overall  & 0.967 & 0.001 & 0.001 & 0.005 & 0.001 & 0.080 & 0.001 & 0.005 & 0.001 & 0.001 \\
         & (-0.38) & (0.68) & (0.76) & (0.40) & (1.00) & (0.24) & (1.00) & (0.48) & (1.00) & (1.00) \\\hline
Summ - Eval  & 0.652 & 0.065 & 0.652 & 0.813 & 0.065 & 0.500 & 0.001 & 0.313 & 0.001 & 0.001 \\
         & (-0.20) & (0.36) & (-0.08) & (-0.24) & (0.36) & (0.08) & (1.00) & (0.02) & (1.00) & (1.00) \\\hline
TL;DR      & 0.019 & 0.116 & 0.053 & 0.313 & 0.001 & 0.019 & 0.001 & 0.116 & 0.001 & 0.001 \\
           & (0.52) & (0.10) & (0.38) & (0.14) & (0.94) & (0.38) & (1.00) & (0.24) & (1.00) & (1.00) \\\hline
UniSum - Eval   & 0.461 & 0.053 & 0.461 & 0.784 & 0.461 & 0.188 & 0.001 & 0.007 & 0.001 & 0.001 \\
              & (0.02) & (0.32) & (0.04) & (-0.20) & (0.08) & (0.22) & (1.00) & (0.24) & (1.00) & (1.00) \\\hline
Dial - Summ - Eval  & 0.001 & 0.024 & 0.935 & 0.839 & 0.188 & 0.348 & 0.001 & 0.014 & 0.001 & 0.001 \\
                & (0.86) & (0.50) & (-0.44) & (-0.38) & (0.20) & (0.06) & (1.00) & (0.36) & (1.00) & (1.00) \\\hline

\multicolumn{11}{c}{\textbf{Relevance}} \\
\hline
Overall   & 0.001 & 0.001 & 0.001 & 0.001 & 0.001 & 0.002 & 0.001 & 0.002 & 0.001 & 0.001 \\
           & (1.00) & (1.00) & (1.00) & (1.00) & (1.00) & (0.72) & (1.00) & (0.82) & (1.00) & (1.00) \\\hline
Dial - Summ - Eval   & 0.001 & 0.001 & 0.001 & 0.001 & 0.065 & 0.216 & 0.001 & 0.001 & 0.001 & 0.001 \\
              & (1.00) & (1.00) & (0.98) & (1.00) & (0.40) & (0.06) & (1.00) & (0.98) & (1.00) & (1.00) \\\hline
Opin - Summ - Eval  & 0.001 & 0.001 & 0.001 & 0.001 & 0.001 & 0.001 & 0.001 & 0.001 & 0.001 & 0.001 \\
                & (0.94) & (0.84) & (0.94) & (0.94) & (0.78) & (0.68) & (1.00) & (0.42) & (1.00) & (1.00) \\\hline
Summ - Eval   & 0.024 & 0.053 & 0.001 & 0.053 & 0.007 & 0.065 & 0.001 & 0.188 & 0.001 & 0.001 \\
            & (0.40) & (0.26) & (0.48) & (0.46) & (0.50) & (0.18) & (1.00) & (0.18) & (1.00) & (1.00) \\\hline
UniSum - Eval   & 0.577 & 0.042 & 0.065 & 0.615 & 0.001 & 0.003 & 0.001 & 0.313 & 0.001 & 0.001 \\
              & (-0.04) & (0.24) & (0.18) & (0.00) & (0.64) & (0.26) & (0.94) & (0.10) & (0.98) & (0.64) \\\hline

\end{longtable}
\end{scriptsize}


\begin{scriptsize}\renewcommand{\arraystretch}{1.5}
\begin{longtable}{|>{\raggedright\arraybackslash}p{1.8cm}>{\raggedright\arraybackslash}p{2.0cm}>{\raggedright\arraybackslash}p{2.0cm}>{\raggedright\arraybackslash}p{2.0cm}>{\raggedright\arraybackslash}p{2.0cm}|}\caption{RAG: p-value of Wilcoxon test for the Tau difference between Jury-On-Demand and Static Jury baselines. Numbers in paranthesis () are effect size Cliff's delta.}\label{tab:sigtest_rag_static}\\
\hline
   \textbf{Data}  & \textbf{Static Jury (Average-All)} & \textbf{Static Jury (Average-TopK)} & \textbf{Static Jury (Weighted-Regression)} & \textbf{Static Jury (Weighted-Tau)}\\
 \hline
\endfirsthead

\hline
  \textbf{Data}  & \textbf{Static Jury (Average-All)} & \textbf{Static Jury (Average-TopK)} & \textbf{Static Jury (Weighted-Regression)} & \textbf{Static Jury (Weighted-Tau)}\\
 \hline
\endhead

\multicolumn{5}{c}{\textbf{Completeness}} \\
\hline

 Overall      &   0.001 &   0.001 &   0.001 &   0.001 \\
              &  (1.00) &   (1.00) &   (1.00) &   (1.00)  \\\hline
 ALCE         &   0.001 &   0.001 &   0.003 &   0.001 \\
               &   (0.58) &   (0.96) &  (0.78) &   (0.94)  \\\hline
 ASQA         &  0.001 &   0.001 &   0.001 &   0.001  \\
              &   (0.96) &   (1.00) &  (1.00) &   (1.00)   \\\hline
 QASPER   &  0.188 &   0.001 &   0.007 &   0.005  \\
               &   (0.24) &   (0.92) &   (0.66)  &   (0.84) \\\hline

\multicolumn{5}{c}{\textbf{Groundedness}} \\
\hline
          Overall      &   0.001 &   0.001 &   0.001 &   0.001 \\
               &   (1.00) &   (1.00) &   (0.82) &   (1.00) \\\hline
 CAQA        &   0.001 &   0.001 &   0.001 &   0.001  \\
               &   (1.00) &   (0.98) &   (0.78) &   (1.00) \\\hline
 Halu - Eval     &   0.002 &   0.001 &   0.005 &   0.001   \\
                &    (0.74) &   (0.48) &   (0.52) &   (1.00)  \\\hline
 Rag - Truth     &   0.001 &   0.001 &   0.010 &   0.001 \\
                &    (0.44) &   (1.00) &   (0.32) &   (1.00)  \\\hline

\multicolumn{5}{c}{\textbf{Relevance}} \\
\hline
  Overall      &   0.001 &   0.001 &   0.003 &   0.001  \\
              &    (0.80) &   (1.00) &   (0.90) &   (1.00)  \\\hline
 ALCE         &   0.001 &   0.001 &   0.066 &   0.001  \\
               &    (0.36) &   (1.00) &   (0.30) &   (1.00) \\\hline
 Hotpot - QA     &   0.001 &   0.003 &   0.001 &   0.001   \\
                &    (0.92) &   (0.74) &   (1.00) &   (0.86) \\\hline
 MS MARCO     &   0.001 &   0.138 &   0.003 &   0.001  \\
                &    (0.84) &   (0.20) &   (0.28) &   (0.36)  \\\hline

\end{longtable}
\end{scriptsize}


\begin{scriptsize}\renewcommand{\arraystretch}{1.5}
\begin{longtable}{|>{\raggedright\arraybackslash}p{0.8cm}>{\raggedright\arraybackslash}p{0.7cm}>{\raggedright\arraybackslash}p{0.7cm}>{\raggedright\arraybackslash}p{0.7cm}>{\raggedright\arraybackslash}p{0.7cm}>{\raggedright\arraybackslash}p{0.7cm}>{\raggedright\arraybackslash}p{0.7cm}>{\raggedright\arraybackslash}p{0.7cm}>{\raggedright\arraybackslash}p{0.7cm}>{\raggedright\arraybackslash}p{0.7cm}>{\raggedright\arraybackslash}p{0.7cm}|}\caption{RAG: p-value of Wilcoxon test for the Tau difference between Jury-On-Demand and single judge. Numbers in paranthesis () are effect size Cliff's delta.}\label{tab:sigtest_rag_judge}\\
\hline
  \textbf{Data}  & \textbf{Claude 3.7} & \textbf{Gemn. 2.0 Flash} & \textbf{Gemn. 2.5 Flash} & \textbf{Gemn. 2.5 Pro} & \textbf{GPT-OSS-20B} & \textbf{GPT-OSS-120B} & \textbf{LL-3.2} & \textbf{Gem\newline ma3} & \textbf{Phi4} & \textbf{Deep\newline Seek\newline R1}\\
 \hline
\endfirsthead

\hline
 \textbf{Data}  & \textbf{Claude 3.7} & \textbf{Gemn. 2.0 Flash} & \textbf{Gemn. 2.5 Flash} & \textbf{Gemn. 2.5 Pro} & \textbf{GPT-OSS-20B} & \textbf{GPT-OSS-120B} & \textbf{LL-3.2} & \textbf{Gem\newline ma3} & \textbf{Phi4} & \textbf{Deep\newline Seek\newline R1}\\
 \hline
\endhead

\multicolumn{11}{c}{\textbf{Completeness}} \\
\hline

 Overall      &   0.001 &   0.001 &   0.001 &   0.001 &   0.001 &  0.001 &   0.001 &   0.001 &  0.001 &  0.001 \\
              &   (1.00) &  (1.00) &   (1.00) &  (1.00) &   (1.00) &   (1.00) &   (1.00) &   (1.00) &   (1.00) &   (1.00) \\\hline
 ALCE         &   0.001 &   0.001 &   0.001 &   0.001 &   0.001 &   0.001  &   0.001 &   0.001 &   0.001  &  0.001 \\
               &   (0.56) &   (0.70) &  (0.80) &   (0.76) &   (0.54)  &   (0.70)  &   (1.00) &   (0.82) &   (1.00) &   (0.88) \\\hline
 ASQA         &  0.001 &   0.001 &   0.001 &   0.001 &   0.001 &  0.001 &   0.001 &   0.001 &  0.001 &  0.001 \\
              &   (0.98) &   (0.98) &   (1.00) &   (1.00) &   (1.00) &   (0.96) &   (1.00) &   (0.98) &   (1.00)  &   (1.00) \\\hline
 QASPER   &   0.005 &   0.001 &   0.246  &   0.005 &   0.053 &   0.216 &   0.001 &   0.002 &   0.001 &  0.001 \\
               &   (0.78) &   (0.96) &   (0.16)  &   (0.64) &   (0.30)  &   (0.10) &   (1.00) &   (0.64) &   (1.00) &   (0.92) \\\hline

\multicolumn{11}{c}{\textbf{Groundedness}} \\
\hline
          Overall      &   0.001 &   0.001 &   0.001 &   0.001  &   0.001 &   0.001 &   0.001 &  0.001 &   0.001 &   0.001  \\
               &   (1.00) &   (1.00) &   (0.96) &   (1.00) &   (0.96) &   (0.92) &   (1.00) &   (1.00) &   (1.00) &  (1.00) \\\hline
 CAQA        &   0.001 &   0.001 &   0.001 &   0.001 &   0.001 &   0.001 &   0.001 &   0.001  &   0.001  &   0.001  \\
               &   (1.00) &   (0.96) &   (1.00) &   (1.00) &   (1.00) &   (0.92) &   (1.00) &   (1.00) &   (1.00) &  (1.00) \\\hline
 Halu - Eval     &   0.001 &   0.001 &   0.014 &   0.001 &   0.024 &   0.461 &   0.001 &   0.001 &   0.001   &   0.001  \\
                &    (1.00) &   (1.00) &   (0.42) &   (0.64) &   (0.28) &   (-0.06) &   (1.00) &   (1.00) &   (1.00) &  (1.00) \\\hline
 Rag - Truth     &   0.001 &   0.001 &   0.001 &   0.001 &  0.001 &   0.001 &   0.001 &   0.001 &   0.001 &   0.001  \\
                &    (1.00) &   (1.00) &   (0.24) &   (0.80) &   (0.72) &   (0.62) &   (1.00) &   (1.00) &   (1.00) &  (1.00) \\\hline

\multicolumn{11}{c}{\textbf{Relevance}} \\
\hline
  Overall      &   0.001 &   0.001 &   0.001 &   0.001  &   0.001 &   0.001 &   0.001 &  0.001 &   0.001  &   0.001  \\
              &    (1.00) &   (1.00) &   (0.98) &   (1.00) &   (1.00) &   (0.98) &   (1.00) &   (1.00) &   (1.00)  &  (1.00) \\\hline
 ALCE         &   0.001 &   0.001 &   0.001 &   0.001  &   0.001 &   0.001 &   0.001 &  0.001 &   0.001  &   0.001  \\
               &    (1.00) &   (0.96) &   (0.62) &   (0.56) &   (0.44) &   (0.48) &   (1.00) &   (1.00) &   (1.00) &  (1.00) \\\hline
 Hotpot - QA     &   0.001 &   0.001 &   0.001 &   0.002  &   0.001 &   0.001 &   0.001 &  0.001 &   0.001  &   0.001  \\
                &    (0.82) &   (1.00) &   (0.34) &   (0.56) &   (0.64) &   (0.22) &   (1.00) &   (1.00) &   (1.00)  &  (1.00) \\\hline
 MS MARCO     &   0.001 &   0.001 &   0.001 &   0.001  &   0.001 &   0.001 &   0.001 &  0.001 &   0.001 &   0.001  \\
                &    (0.72) &   (0.66) &   (0.52) &   (0.34) &   (0.80) &   (0.52) &   (1.00) &   (1.00) &   (1.00) &  (1.00) \\\hline

\end{longtable}
\end{scriptsize}

\subsection{Analyzing the Interaction Between Judges, Tasks, and Data Attributes}\label{sec:full_res_2}

\begin{figure}[htbp]
    \centering
    \begin{subfigure}[b]{0.49\textwidth}
        \includegraphics[width=\textwidth]{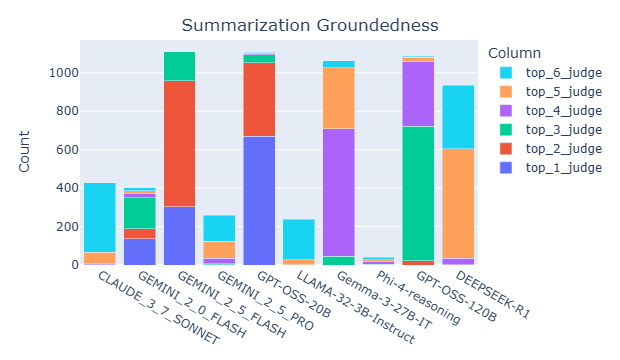}
    \end{subfigure}
    \hfill
    \begin{subfigure}[b]{0.49\textwidth}
        \includegraphics[width=\textwidth]{media/image_dg08_grd.png}
    \end{subfigure}

    \vspace{0.5cm}

    \begin{subfigure}[b]{0.49\textwidth}
        \includegraphics[width=\textwidth]{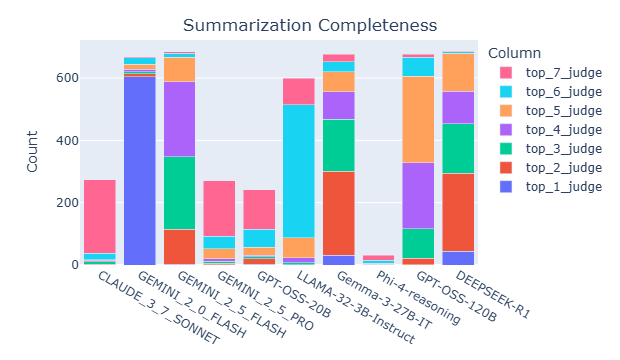}
    \end{subfigure}
    \hfill
    \begin{subfigure}[b]{0.49\textwidth}
        \includegraphics[width=\textwidth]{media/image_dg08_cmpl.png}
    \end{subfigure}

    \vspace{0.5cm}

    \begin{subfigure}[b]{0.49\textwidth}
        \includegraphics[width=\textwidth]{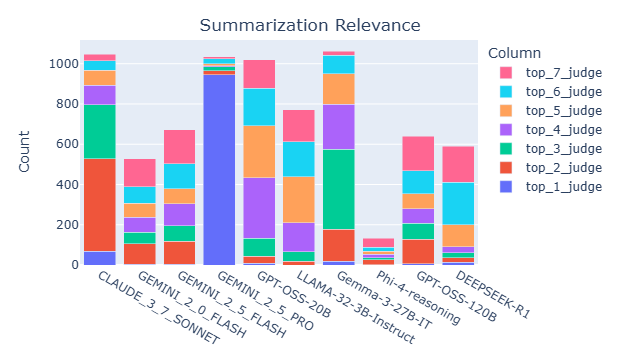}
    \end{subfigure}
    \hfill
    \begin{subfigure}[b]{0.49\textwidth}
        \includegraphics[width=\textwidth]{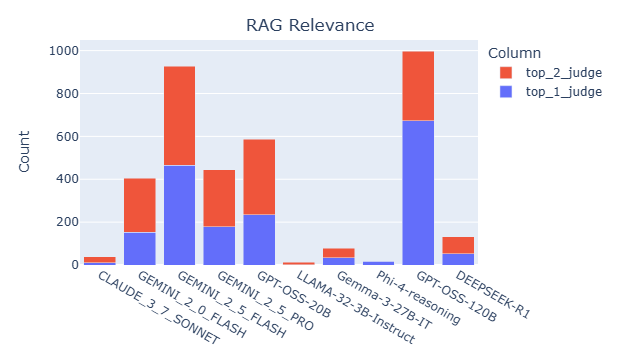}
    \end{subfigure}

    \caption{Selection frequency of the judge in the jury. Top k judge means that the judge has the k-th highest reliability score in the jury.  Summarization juries tend to incorporate a more diverse set of judges compared to those used in RAG tasks. For RAG Claude 3.7 Sonnet and DeepSeek R1 are frequently selected for completeness evaluation but are rarely chosen for groundedness. In contrast, Gemini 2.5 Flash is commonly selected for groundedness but appears less frequently in completeness evaluations. GPT OSS 20B and GPT OSS 120B are consistently selected across both metrics.}
    \label{fig:sel_freq_judge_jury}
\end{figure}

Fig. \ref{fig:sel_freq_judge_jury} summarizes the selection frequency of each judge, revealing that summarization juries tend to include a broader set of judges compared to those used in RAG tasks. Within RAG, Claude 3.7 Sonnet and DeepSeek R1 are frequently selected for completeness evaluation but are rarely chosen for groundedness. In contrast, Gemini 2.5 Flash is commonly selected for groundedness but appears less frequently in completeness evaluations. GPT OSS 20B and GPT OSS 120B are consistently selected across both metrics.

\begin{figure}[htbp]
\centering
\begin{minipage}[b]{0.48\textwidth}
  \centering
  \includegraphics[width=\textwidth]{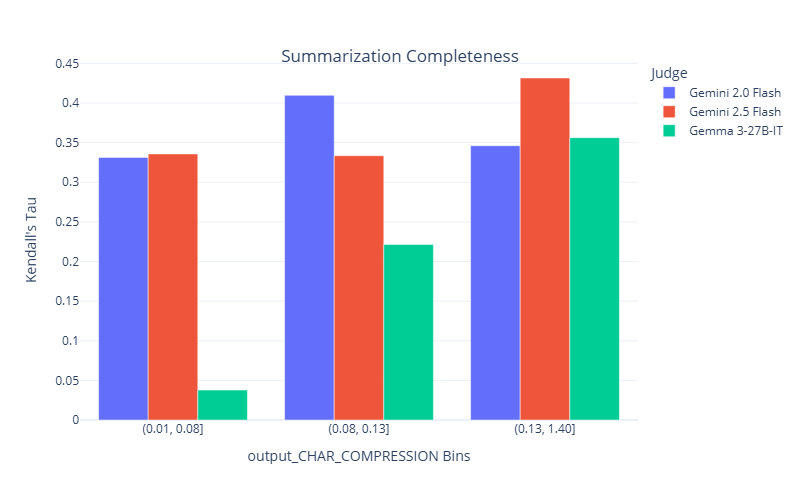}
\end{minipage}
\hfill
\begin{minipage}[b]{0.48\textwidth}
  \centering
  \includegraphics[width=\textwidth]{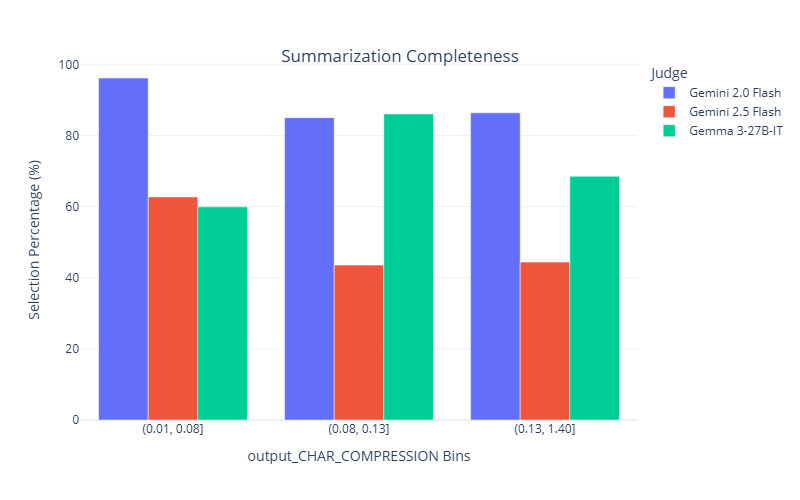}
\end{minipage}
\caption{Summarization Completeness analysis by character compression ratio. (Left) Kendall's Tau correlation for three single judges across low, medium, and high compression ratio bins. (Right) The selection percentage of three judges in the final dynamic jury for data points within each bin. Performance generally degrades at lower compression ratio, with Gemma 3-27B-IT failing significantly in this region. The jury selection percentage correctly mirrors this, heavily selecting Gemini 2.0 Flash which is the most reliable in that bin.}
\label{fig:bin_summ}
\end{figure}

To further explore how data properties influence model performance and judge selection, we examine Kendall's tau and selection patterns across different bins of key attributes. Following Section \ref{sec:diag_property}. We focus on summarization completeness task and property compression ratio. Fig. \ref{fig:bin_summ} compares judge performance across low, medium, and high compression ratio bins. Performance improves as the compression ratio increases, likely because judges find it more difficult to identify incomplete summaries, which tend to have lower compression ratios. Fig. \ref{fig:bin_summ} shows that Gemma performs particularly poorly when the compression ratio is low. Upon reviewing the data, Gemma’s scores, and its explanations, we found that it sometimes struggles to distinguish between the source context and the summary. As a result, it incorrectly assigns a score of 2, interpreting the context as part of the summary. Examples illustrating this issue are provided in the Section \ref{sec:full_res_3_example}. Stronger models—such as the Gemini series, GPT models, and Claude 3.7 Sonnet—do not exhibit this issue and are able to correctly identify missing content in the summaries.

Jury selection, shown in the right plot, is generally consistent with judge performance across the bins. In the low compression bin where Gemma fails, it is selected least often. In contrast, Gemini 2.0 Flash has the highest performance in this bin and is selected in nearly all juries, showing strong alignment. However, the selection mechanism again proves more sophisticated than just selecting the top-ranked judge. In the high compression ratio bin, Gemini 2.5 Flash (red) achieves the highest Kendall's Tau. Yet, Gemini 2.0 Flash (blue), which also performs well, is selected more frequently. This demonstrates that the reliability model identifies multiple judges as reliable in this context and dynamically constructs juries based on this broader reliability assessment rather than overfitting to a single ``best'' judge for the bin.

These findings reinforce the importance of constructing dynamic juries that adapt to specific data characteristics, and demonstrate the potential of predicting judge reliability based on interpretable data properties.

\pagebreak
Tables \ref{fig: top_ftrs_judge_13}-\ref{fig: top_ftrs_judge_45} present the top five most important features for each judge’s XGBoost model, as determined by permutation feature importance \citep{fisher2019all}, using the summarization groundedness task as an illustrative example. The results show substantial variation in the top-ranked features across different judges, suggesting that each judge’s reliability is influenced by distinct data properties.

\begin{figure}[htbp]
    \centering
    \begin{subfigure}[b]{0.49\textwidth}
        \includegraphics[width=\textwidth]{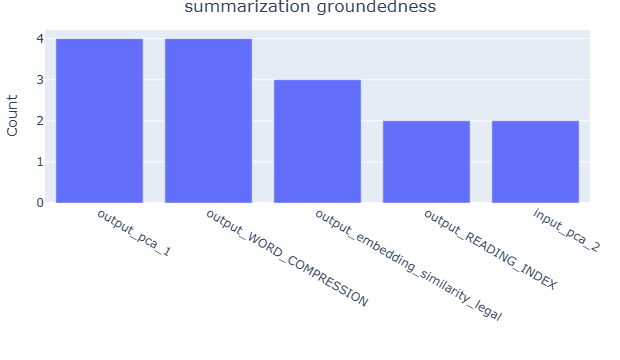}
    \end{subfigure}
    \hfill
    \begin{subfigure}[b]{0.49\textwidth}
        \includegraphics[width=\textwidth]{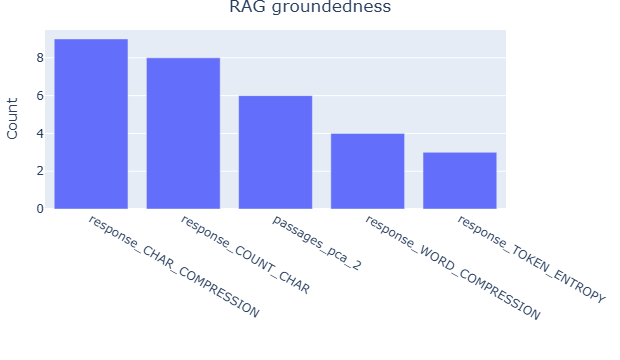}
    \end{subfigure}

    \vspace{0.5cm}

    \begin{subfigure}[b]{0.49\textwidth}
        \includegraphics[width=\textwidth]{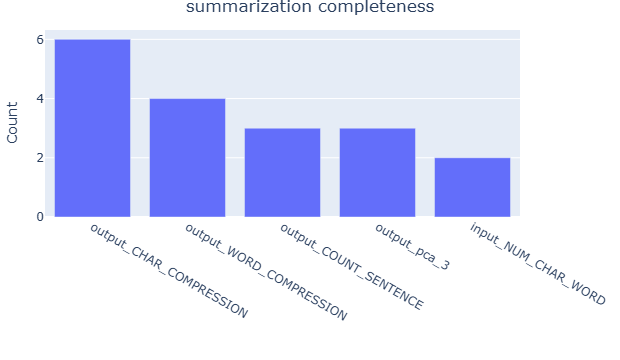}
    \end{subfigure}
    \hfill
    \begin{subfigure}[b]{0.49\textwidth}
        \includegraphics[width=\textwidth]{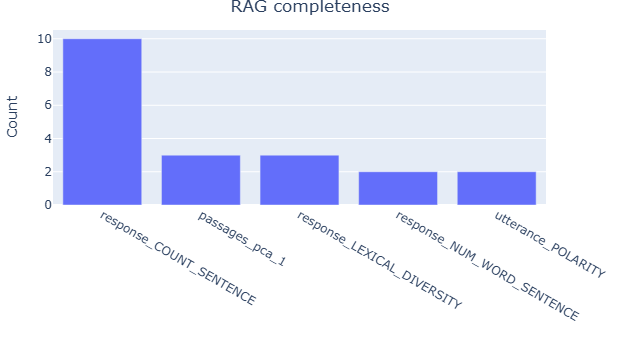}
    \end{subfigure}

    \vspace{0.5cm}

    \begin{subfigure}[b]{0.49\textwidth}
        \includegraphics[width=\textwidth]{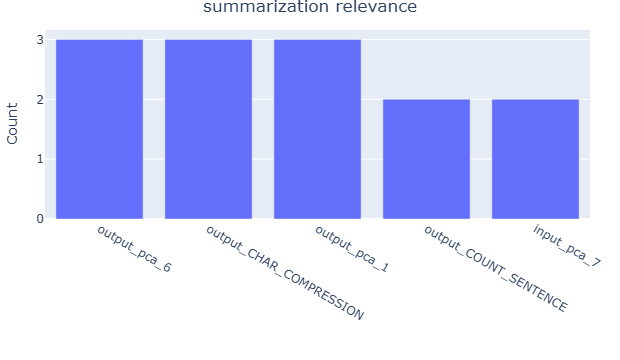}
    \end{subfigure}
    \hfill
    \begin{subfigure}[b]{0.49\textwidth}
        \includegraphics[width=\textwidth]{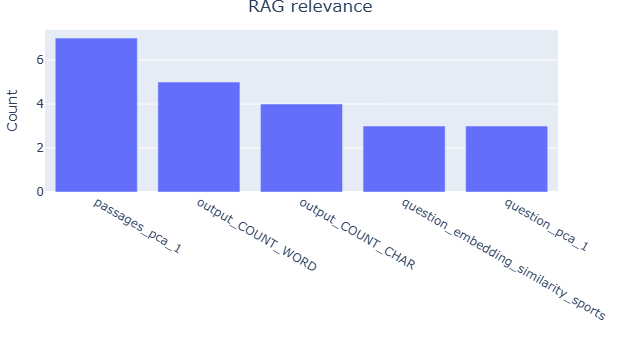}
    \end{subfigure}
    \caption{Aggregating the top five features that frequently appear across tasks. Character count is more prominent in RAG tasks, while compression ratio and embedding-related features such as PCA components and embedding similarity are more influential in summarization tasks.}
    \label{fig:blue_bar_plot}
\end{figure}

Fig. \ref{fig:blue_bar_plot} aggregates the top five features that frequently appear across tasks, revealing clear task-specific patterns. For instance, character count is more prominent in RAG tasks, while compression ratio and embedding-related features such as PCA components and embedding similarity are more influential in summarization tasks. These findings align with the ablation analysis in Appendix \ref{appendix:ablation}, which shows that removing embedding features leads to a greater performance drop in the summarization task compared to RAG. For RAG, removing text size-related features results in a larger decline than removing embedding features.

These observations imply that evaluation reliability is task-dependent, and further demonstrate that our approach effectively links data characteristics to judge reliability, enabling more informed and adaptive jury construction across diverse evaluation scenarios.

\begin{table}[htbp]
\caption{The top 3 most important features for each judge’s XGBoost model from summarization groundedness. The results show substantial variation in the top-ranked features across different judges.}
\label{fig: top_ftrs_judge_13}
\centering
\renewcommand{\arraystretch}{1.5}\tiny
\begin{tabular}{>{\raggedright\arraybackslash}p{0.2\textwidth} >{\raggedright\arraybackslash}p{0.2\textwidth} >{\raggedright\arraybackslash}p{0.2\textwidth} >{\raggedright\arraybackslash}p{0.2\textwidth} }
\toprule
\textbf{Judge} & \textbf{Feature 1} & \textbf{Feature 2} & \textbf{Feature 3} \\
\midrule
CLAUDE 3 .7 SONNET & input \_ embedding \_ similarity \_ politics & input \_ COUNT \_ WORD & input \_SUBJECTIVITY\\\hline
GEMINI 2.0 FLASH & input\_pca\_9 & output\_pca\_9 & output \_pca \_7  \\\hline
GEMINI 2.5 FLASH & input\_pca\_9 & output \_embedding \_similarity \_legal & output\_pca\_1 \\\hline
GEMINI 2.5 PRO & input \_embedding \_similarity \_financemarket & output \_embedding \_similarity \_financebank & output \_CHAR \_COMPRESSION  \\\hline
GPT-OSS-20B & output \_pca \_1 & output \_WORD \_COMPRESSION & output \_pca \_2  \\\hline
LLAMA-32-3B-Instruct & output \_READING \_INDEX & input \_embedding \_similarity \_media & input \_READING \_INDEX  \\\hline
Gemma-3-27B-IT & input \_embedding \_similarity \_sports & output \_SENTENCE \_SIMILARITY & output \_embedding \_similarity \_media  \\\hline
Phi-4-reasoning & input \_DIFFICULT \_WORD & output \_NUM \_WORD \_SENTENCE & output \_LEXICAL \_DIVERSITY \\\hline
GPT-OSS-120B & input \_embedding \_similarity \_financemarket & output \_embedding \_similarity \_financemarket & output \_CHAR \_COMPRESSION \\\hline
DEEPSEEK-R1 & input\_pca\_6 & output \_embedding \_similarity \_legal & output \_embedding \_similarity \_financemarket  \\
\bottomrule
\end{tabular}
\end{table}

\begin{table}[htbp]
\caption{The 4th and 5th most important features for each judge’s XGBoost model from summarization groundedness. The results show substantial variation in the top-ranked features across different judges.}
\label{fig: top_ftrs_judge_45}
\centering
\renewcommand{\arraystretch}{1.2}\tiny
\begin{tabular}{>{\raggedright\arraybackslash}p{0.2\textwidth} >{\raggedright\arraybackslash}p{0.25\textwidth} >{\raggedright\arraybackslash}p{0.25\textwidth}}
\toprule
\textbf{Judge}  & \textbf{Feature 4} & \textbf{Feature 5} \\
\midrule
CLAUDE 3 .7 SONNET & output \_DIFFICULT \_WORD & output \_pca \_8 \\\hline
GEMINI 2.0 FLASH  & input\_pca\_7 & output \_pca \_3 \\\hline
GEMINI 2.5 FLASH  & input \_pca \_1 & output \_WORD \_COMPRESSION \\\hline
GEMINI 2.5 PRO & output \_pca \_3 & output \_pca \_1 \\\hline
GPT-OSS-20B  & output \_pca \_9 & output \_embedding \_similarity \_business \\\hline
LLAMA-32-3B-Instruct  & output \_ COUNT \_ WORD & output \_pca \_1 \\\hline
Gemma-3-27B-IT  & output \_COREFERENCE \_CHAIN & input \_SEMANTIC \_AMBIGUITY \\\hline
Phi-4-reasoning & output \_pca \_5 & output \_READING \_INDEX \\\hline
GPT-OSS-120B  & output \_pca \_3 & output \_pca \_1 \\\hline
DEEPSEEK-R1  & output \_SENTENCE \_SIMILARITY & input \_NAMED \_ENTITIE \\
\bottomrule
\end{tabular}
\end{table}

\subsection{Human-Reviewed Examples}\label{sec:full_res_3_example}

\begin{figure}[h]
\centering
\includegraphics[width=0.9\textwidth]{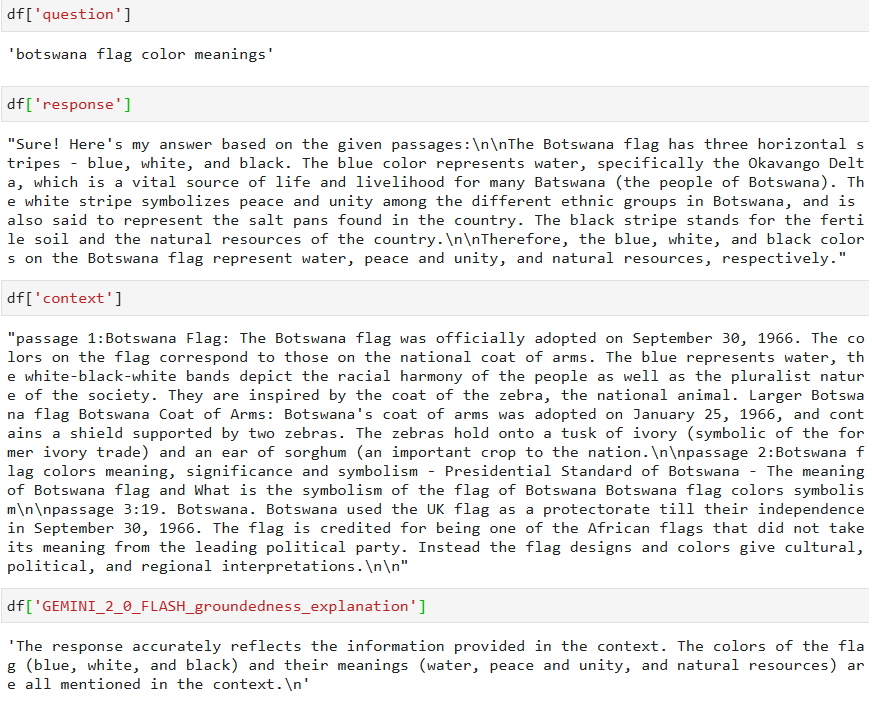}
\caption{Gemini 2.0 Flash fails to identify ungrounded content in the response—for instance, Okavango Delta, which is not mentioned in the cited context.}
\label{fig:example_review_rag}
\end{figure}

\begin{figure}[h]
\centering
\includegraphics[width=0.9\textwidth]{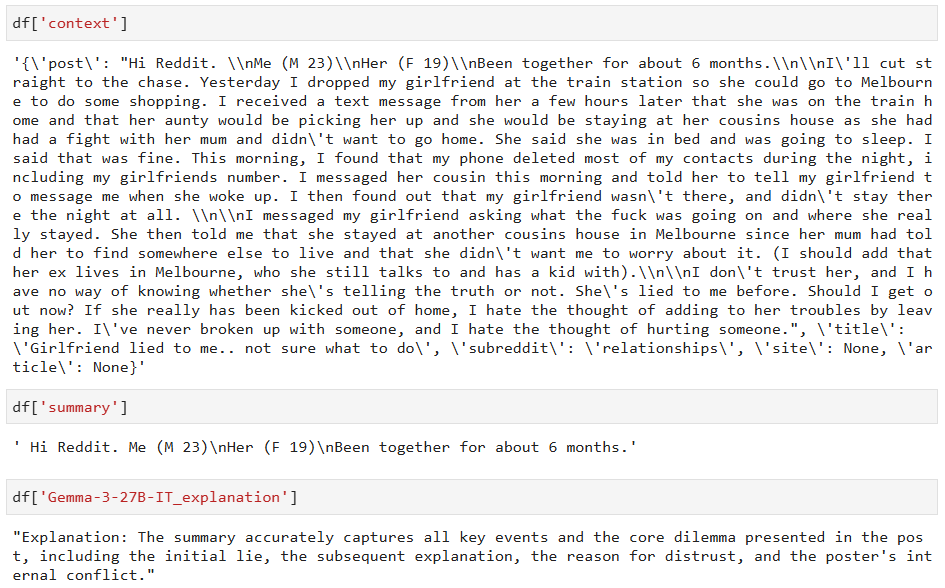}
\caption{Gemma 3 27B IT fails to assess the completeness of a summary. The summary merely repeats the first sentence of the post and omits key details, yet Gemma incorrectly considers it complete and explains that it covers all key events.}
\label{fig:example_review_summ}
\end{figure}

In this section, we present examples where specific judges fail to evaluate correctly. For illustration, we select one example and one judge from each of the two tasks: RAG groundedness and summarization completeness. Fig. \ref{fig:example_review_rag} shows how Gemini 2.0 Flash evaluates groundedness in the RAG task. It fails to identify ungrounded content in the response—for instance, Okavango Delta, which is not mentioned in the cited context. Fig. \ref{fig:example_review_summ} illustrates how Gemma 3 27B IT fails to assess the completeness of a summary. The summary merely repeats the first sentence of the post and omits key details, yet Gemma incorrectly considers it complete and explains that it covers all key events.

\subsection{Explanations on Embedding PCA-Related Features}\label{sec:full_res_4_pca}

\begin{figure}[htbp]
\centering
\begin{minipage}[b]{0.48\textwidth}
  \centering
  \includegraphics[width=\textwidth]{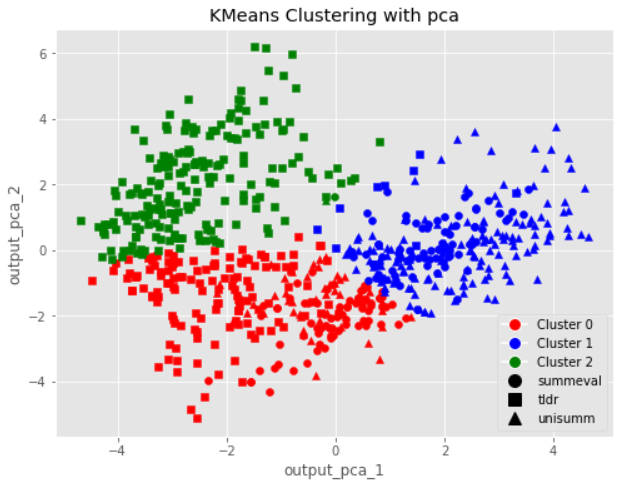}
\end{minipage}
\hfill
\begin{minipage}[b]{0.48\textwidth}
  \centering
  \includegraphics[width=\textwidth]{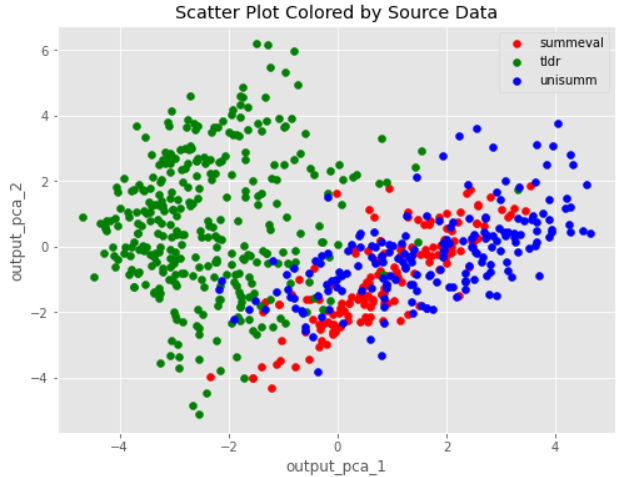}
\end{minipage}
\caption{ The left figure shows the data distribution after applying K-means clustering based on the first two principal components (PCA 1 and PCA 2) of the embedding features. The right figure displays the same data distribution, colored by source dataset. It is evident that under PCA 1 and PCA 2, the TLDR dataset is clearly separated from the other two datasets. }
\label{fig:pca_data}
\end{figure}

In this section, we provide insights into the embedding PCA-related features. For illustration, we use the summarization completeness task as an example. Fig. \ref{fig:pca_data} (left) shows the data distribution after applying K-means clustering based on the first two principal components (PCA 1 and PCA 2) of the embedding features. Fig. \ref{fig:pca_data} (right) displays the same data distribution, colored by source dataset. It is evident that under PCA 1 and PCA 2, the TLDR dataset is clearly separated from the other two datasets, indicating that these embedding features capture meaningful dataset-level information. A likely explanation for this separation lies in differences in topic and writing style. TLDR samples originate from Reddit posts, which typically focus on personal or emotional dilemmas and are written in a less formal style. In contrast, source contexts from SummEval and UniSumm consist of more formally written news articles and reports. This observation is further supported by topic modeling using Latent Dirichlet Allocation (LDA) \citep{blei2003latent}. Below, we summarize the top words and inferred topics for each dataset:
\begin{itemize}
\item SummEval: Top words include club, hull, liverpool, time, and claim, suggesting topics related to sports and crime.
  \item UniSumm: Top words include text, rate, city, and officer, indicating topics such as technical instructions and city-related events.
  \item TLDR: Top words include want, boyfriend, girlfriend, love, and problem, reflecting themes of romantic desires and relationship issues.
\end{itemize}

\pagebreak
\subsection{Analysis of Jury Failure in Evaluation Tasks}\label{appendix:jury_fail_bin}
\begin{figure}[htbp]
    \centering
    \begin{subfigure}[b]{0.49\textwidth}
        \includegraphics[width=\textwidth]{media/RAG_groundedness_response_TOKEN_ENTROPY_bin.png}
    \end{subfigure}
    \hfill
    \begin{subfigure}[b]{0.49\textwidth}
        \includegraphics[width=\textwidth]{media/RAG_groundedness_response_COUNT_CHAR_bin.png}
    \end{subfigure}

    \vspace{0.5cm}

    \begin{subfigure}[b]{0.49\textwidth}
        \includegraphics[width=\textwidth]{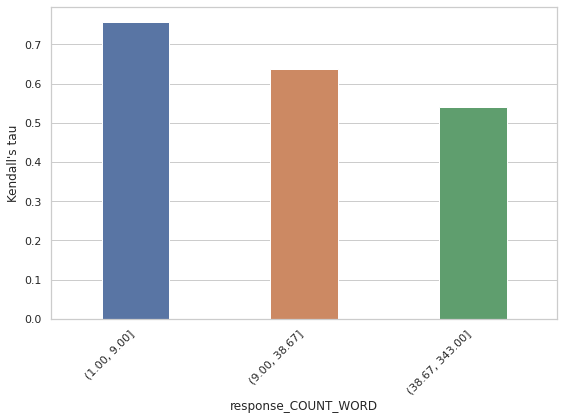}
    \end{subfigure}
    \hfill
    \begin{subfigure}[b]{0.49\textwidth}
        \includegraphics[width=\textwidth]{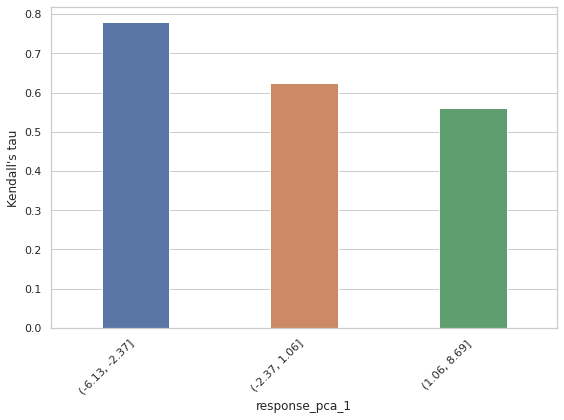}
    \end{subfigure}

    \vspace{0.5cm}

    \begin{subfigure}[b]{0.49\textwidth}
        \includegraphics[width=\textwidth]{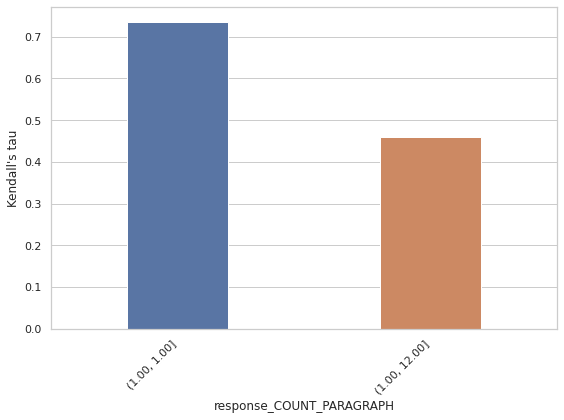}
    \end{subfigure}
    \hfill
    \caption{Jury performance across segments of top features indentifed in the RAG goundedness task. Jury has better outcomes when the generated text is shorter or less complex.}
    \label{fig:jury_rag_bin_all}
\end{figure}

Figure \ref{fig:jury_rag_bin_all} presents the complete results of binning analyses of top features for the RAG groundedness task. The results reveal a clear trend: as the length or complexity of generated text increase, the jury's ability to reliably assess groundedness decreases. It is hard to interpret the meaning of PCA features of generated text but a clear pattern is also shown.

\pagebreak
\section{Human Evaluation Consistency}\label{appendix:comp_human_eval}

We examine the consistency of human evaluations and compare them with the performance of our jury model. Both the SummEval and DialSumm datasets include three human annotators per evaluation dimension. Tables \ref{tab:correlation_matrix_1} to \ref{tab:correlation_matrix_4} report Kendall’s tau correlations between each individual annotation and the average of the three scores, which serves as the reference in our analysis. The top-left cell in each table presents the Kendall’s tau of the jury model. Our findings reveal that inter-annotator agreement is generally low, highlighting notable discrepancies among human judgments. While Kendall’s tau increases when comparing individual annotator scores to the average, substantial variation remains. For instance, in the SummEval Relevance dimension, the highest tau between a human annotator and the average is 0.875, whereas the lowest is only 0.411. Although the jury model rarely surpasses the best-performing annotator, it consistently outperforms the weaker ones. This suggests that the jury model offers a reliable and robust alternative to individual human evaluations.

\begin{table}[h!]
\caption{Kendall’s tau for human annotations – SummEval Groundedness}
\label{tab:correlation_matrix_1}
\centering
\begin{tabular}{|l|c|c|c|c|}
\hline
 \multicolumn{1}{|c|}{0.576} & \textbf{Average score} & \textbf{Annotator 1} & \textbf{Annotator 2} & \textbf{Annotator 3} \\
\hline
\textbf{Average score} & 1 & 0.893 & 0.893 & 0.735 \\
\textbf{Annotator 1} & 0.893 & 1 & 0.748 & 0.793 \\
\textbf{Annotator 2} & 0.893 & 0.748 & 1 & 0.807 \\
\textbf{Annotator 3} & 0.735 & 0.793 & 0.807 & 1 \\
\hline
\end{tabular}
\end{table}

\begin{table}[h!]
\caption{Kendall’s tau for human annotations – SummEval Relevance.}
\label{tab:correlation_matrix_2}
\centering
\begin{tabular}{|c|c|c|c|c|}
\hline
\multicolumn{1}{|c|}{0.696} & \textbf{Average score} & \textbf{Annotator 1} & \textbf{Annotator 2} & \textbf{Annotator 3} \\
\hline
\textbf{Average score} & 1     & 0.667 & 0.875 & 0.411 \\
\textbf{Annotator 1}   & 0.667 & 1     & 0.455 & 0.388 \\
\textbf{Annotator 2}   & 0.875 & 0.455 & 1     & 0.394 \\
\textbf{Annotator 3}   & 0.411 & 0.388 & 0.394 & 1     \\
\hline
\end{tabular}
\end{table}

\begin{table}[h!]
\caption{Kendall’s tau for human annotations – DialSumm Groundedness.}
\label{tab:correlation_matrix_3}
\centering
\begin{tabular}{|c|c|c|c|c|}
\hline
\multicolumn{1}{|c|}{0.699} & \textbf{Average score} & \textbf{Annotator 1} & \textbf{Annotator 2} & \textbf{Annotator 3} \\
\hline
\textbf{Average score} & 1     & 0.647 & 0.712 & 0.679 \\
\textbf{Annotator 1}   & 0.647 & 1     & 0.462 & 0.379 \\
\textbf{Annotator 2}   & 0.712 & 0.463 & 1     & 0.351 \\
\textbf{Annotator 3}   & 0.679 & 0.379 & 0.351 & 1     \\
\hline
\end{tabular}

\end{table}

\begin{table}[h!]
\caption{Kendall’s tau for human annotations – DialSumm Relevance.}
\label{tab:correlation_matrix_4}
\centering
\begin{tabular}{|c|c|c|c|c|}
\hline
\multicolumn{1}{|c|}{0.639} & \textbf{Average score} & \textbf{Annotator 1} & \textbf{Annotator 2} & \textbf{Annotator 3} \\
\hline
\textbf{Average score} & 1     & 0.741 & 0.758 & 0.622 \\
\textbf{Annotator 1}   & 0.741 & 1     & 0.564 & 0.365 \\
\textbf{Annotator 2}   & 0.758 & 0.564 & 1     & 0.344 \\
\textbf{Annotator 3}   & 0.622 & 0.365 & 0.344 & 1     \\
\hline
\end{tabular}
\end{table}

\section{Judge Reliability Prediction Model Performance}\label{appendix:reliability_perf}

Table \ref{tab:auc_judge_rel} presents the AUC scores of ROC curves for each judge reliability model trained using XGBoost on the testing set. The AUC values are relatively consistent across tasks; we include results for summarization completeness and RAG groundedness as representative examples. Result varies across judges and tasks, with most AUCs ranging between 0.63 and 0.78, indicating that the models demonstrate adequate predictive capability for these evaluation tasks.

\begin{table}[htbp]
\centering
\scriptsize
\caption{AUC of ROC for the judge reliability model. Here Gemn. is Gemini, DS is DeepSeek.}
\label{tab:auc_judge_rel}
\begin{tabular}{p{1.7cm}|p{0.7cm}p{0.7cm}p{0.7cm}p{0.7cm}p{0.7cm}p{0.7cm}p{0.7cm}p{0.7cm}p{0.7cm}p{0.7cm}}
\toprule
\textbf{Metric} & \textbf{Claude 3.7 SONNET} & \textbf{Gemn. 2.0 Flash} & \textbf{Gemn. 2.5 Flash} & \textbf{Gemn. 2.5 Pro} & \textbf{GPT-OSS-20B} & \textbf{GPT-OSS-120B} & \textbf{LLAMA -3.2-3B Instruct} & \textbf{Gemma 3-27B-IT} & \textbf{Phi-4-reason - ing}  & \textbf{DS-R1}\\
\midrule
Summ -completeness & 0.63 & 0.67 & 0.72 & 0.66 & 0.76 & 0.75 & 0.63 & 0.63 & 0.62 & 0.68\\ \hline
RAG -groundedness           & 0.70 & 0.78 & 0.67 & 0.68 & 0.76 & 0.73 & 0.61 & 0.77 & 0.75 & 0.73\\
\bottomrule
\end{tabular}
\end{table}

\section{Ablation Study and Model Weakness Analysis}

\subsection{Ablation Study (Data Properties)}\label{appendix:ablation}

To investigate the influence of data property features on jury performance, we conduct ablation studies by selectively removing feature sets during model construction. As illustrative examples, we focus on two tasks: summarization completeness and RAG groundedness. The full categorization of features is provided in Appendix \ref{appendix:list-of-data-features}. Specifically, we remove three groups of features in separate experiments: (1) text size-related features and special word count features (jointly removed due to their high correlation), (2) text complexity features, and (3) embedding-based features. The results are presented in Tables \ref{tab:corr_summ_ablation} and \ref{tab:corr_rag_ablation} for summarization completeness and RAG groundedness, respectively. We observe that the jury model achieves its best performance when all feature sets are included, underscoring the importance of comprehensive feature representation. Although the performance differences are modest, this can be attributed to the internal correlations within each feature category. Additionally, different tasks exhibit varying sensitivity to feature sets. For instance, removing embedding features leads to a greater performance drop in the summarization completeness task than in RAG groundedness, where text size-related features have a more pronounced impact.

\begin{table}[h!]
\caption{Kendall’s tau for jury performance on summarization completeness under feature ablation. The number of judges (k) is indicated for each configuration. The jury using all features achieves the highest performance, particularly on SummEval.}
\label{tab:corr_summ_ablation}
\centering
\scriptsize
\begin{tabular}{|l|p{2cm}|p{2cm}|p{2cm}|p{2cm}|}
\hline
\textbf{Data} & \textbf{Text size + special words removed (k=6)} & \textbf{Text complexity removed (k=5)} & \textbf{Embedding removed (k=6)} & \textbf{All features (k=7)} \\
\hline
Overall      & 0.487 & 0.478 & 0.471 & 0.488 \\
SummEval     & 0.639 & 0.658 & 0.688 & 0.721 \\
TL;DR        & 0.352 & 0.459 & 0.318 & 0.427 \\
UniSumEval   & 0.619 & 0.595 & 0.589 & 0.612 \\
\hline
\end{tabular}
\end{table}

\begin{table}[h!]
\caption{Kendall’s tau for jury performance on RAG grounded under feature ablation. The number of judges (k) is indicated for each configuration. The jury using all features achieves the highest performance, particularly on RagTruth.}
\label{tab:corr_rag_ablation}
\centering
\scriptsize
\begin{tabular}{|l|p{2cm}|p{2cm}|p{2cm}|p{2cm}|}
\hline
\textbf{Data} & \textbf{Text size + special words removed (k=4)} & \textbf{Text complexity removed (k=3)} & \textbf{Embedding removed (k=3)} & \textbf{All features (k=3)} \\
\hline
Overall    & 0.667 & 0.672 & 0.667 & 0.678 \\
CAQA       & 0.659 & 0.659 & 0.676 & 0.651 \\
HaluEval   & 0.761 & 0.759 & 0.766 & 0.798 \\
RagTruth   & 0.558 & 0.596 & 0.575 & 0.576 \\
\hline
\end{tabular}
\end{table}

\pagebreak
\subsection{Ablation Study (Jury Size - $K$)}\label{appendix:ablation_jury_size}

\begin{figure}[htbp]
\centering
\begin{minipage}[b]{0.48\textwidth}
  \centering
  \includegraphics[width=\textwidth]{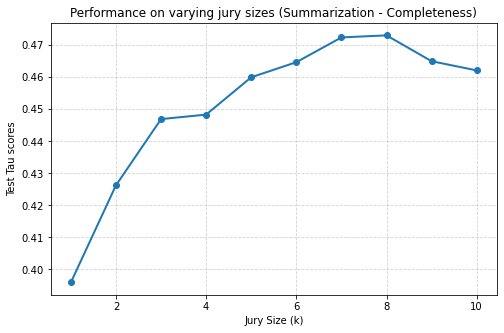}
\end{minipage}
\hfill
\begin{minipage}[b]{0.48\textwidth}
  \centering
  \includegraphics[width=\textwidth]{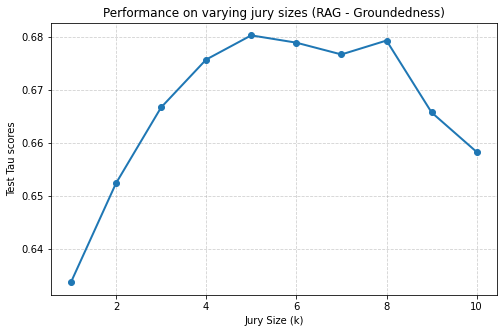}
\end{minipage}
\caption{Test performance with varying jury size.}
\label{fig:ablation_jury_size}
\end{figure}

We conduct experiments to test the effectiveness of varying jury size compared to keeping a fixed value.
We focus on the tasks Summarization-Completeness and RAG-Groundedness for this study. The tolerance level for all trained XGBoost
models is set to 0 which means only the exactly matching scores are considered as correct.
Performance is measured across 10 runs and average is taken for each jury size. Fig. \ref{fig:ablation_jury_size}
shows that the performance varies with jury size.
In the Summarization-Completeness task, performance steadily improves as jury size increases, reaching its peak around a
jury size of 7–8 before slightly declining. This indicates that adding more judges generally strengthens the overall decision
by reducing individual biases, but very large juries introduce diminishing returns and slight degradation. Similarly, in the
RAG-Groundedness task, performance starts low at smaller jury sizes and improves significantly with larger juries, peaking
at 5 and 8 and then tapering off. In both cases, increasing jury size enhances robustness against noisy predictions,
but there is an optimal range beyond which gains flatten or reverse. This shows that tuning jury size per task still
provides significant improvements compared to fixing it, though the optimal size tends to be moderately large for both tasks.

\subsection{Ablation Study (Tolerance levels - $\tau$)}\label{appendix:ablation_tolerance_levels}

We consider two tolerance levels for summarization tasks (0 and 1) and a
single tolerance level (0) for RAG tasks. For RAG tasks valid scores
are 0-2. Thus, a tolerance level of 1 or more would mean that a score of
1 is always considered as correct. For the same reason, we do not
experiment with a tolerance level of 2 or more for summarization tasks
where the valid scores are 1-5.

For this study, we focus on Summarization-Completeness. We experiment with two tolerance
levels: 0 and 1 in the original scale (1-5). After min-max normalization to {[}0, 1{]}, these correspond to 0 and 0.25. The performance of the jury is observed with all the XGBoost models trained either on
tolerance of 0 or 0.25. Table \ref{tab:corr_tol_summ_ablation} summarizes the means and standard
deviations of the Kendall's Tau across the 10 runs and shows the
comparison with tuned tolerance models. We observe that allowing
different tolerance levels across different XGBoost models gives
slightly better performance than a fixed tolerance level across all
models.

Additionally, Table \ref{tab:corr_tol_summ_ablation} further illustrates the importance of tolerance tuning.
While Jury-on-Demand with variable tolerance achieves the best overall performance, the optimal fixed tolerance differs across datasets: TL;DR performs better with a tolerance of 0, whereas UniSumEval favors 0.25.
This variability underscores that no single fixed tolerance can fit all datasets. In practical scenarios,
especially for unseen datasets without human annotations, it is impossible to know the ideal tolerance beforehand.
Therefore, adaptive approaches that allow tolerance to vary across models or instances are crucial for robust generalization.

Finally, the results with fixed tolerance levels are better than the static jury
as we have chosen the best jury size ($K=7$) overall across the runs.

\begin{table}[h!]
\caption{Kendall’s tau for jury performance on Summarization-Completeness under tolerance ablation. The number of judges ($K$) is fixed to 7 (overall best) for the fixed tolerance configurations.}
\label{tab:corr_tol_summ_ablation}
\centering
\scriptsize
\begin{tabular}{|l|p{2cm}|p{2cm}|p{2cm}|p{2cm}|}
\hline
\textbf{Data} & \textbf{Fixed tolerance (0)} & \textbf{Fixed tolerance (0.25)} & \textbf{Jury on Demand (variable tolerance)} & \textbf{Static Jury} \\
\hline

Overall     & 0.47 (0.02) & 0.47 (0.00) & 0.48 (0.03) & 0.44 (0.02) \\
SummEval    & 0.69 (0.05) & 0.75 (0.04) & 0.72 (0.05) & 0.60 (0.06) \\
TL;DR       & 0.38 (0.06) & 0.37 (0.04) & 0.38 (0.08) & 0.40 (0.05) \\
UniSumEval  & 0.63 (0.04) & 0.66 (0.05) & 0.66 (0.04) & 0.59 (0.03) \\
\hline
\end{tabular}
\end{table}

\subsection{Ablation Study (Prompt Variation)}\label{appendix:ablation_prompt_effect}

We analyze the effect of using different prompts with the judges on the overall jury performance. We focus on the task
Summarization-Completeness for our analysis. The performance with the prompt as described in Table \ref{tab:prompt_summarize} is compared against
the performance with a slightly different prompt which omits the list of valid scores.

\begin{figure}[h]
\centering
\includegraphics[width=1.0\textwidth]{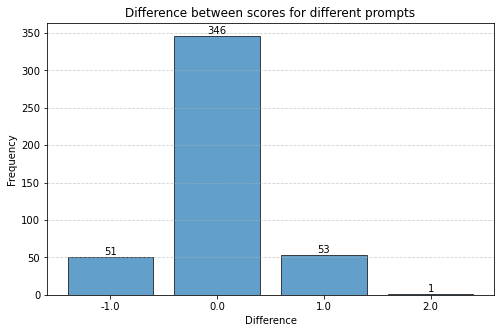}
\caption{Score differences in the results of GEMINI 2.5 PRO judge with two different prompts}
\label{fig:prompt_perturb_score_difference}
\end{figure}

We observe that changing the prompt while maintaining its meaningfulness (the prompt should clearly explain the input
and the expected output), changes the evaluation scores for roughly $30\%$ of the data points by 1 score point
(on average across datasets, judges and metrics). Fig. \ref{fig:prompt_perturb_score_difference} shows the distribution of score differences
for judging the Completeness of summaries using GEMINI 2.5 PRO as the judge.

\begin{figure}[h]
\centering
\includegraphics[width=1.0\textwidth]{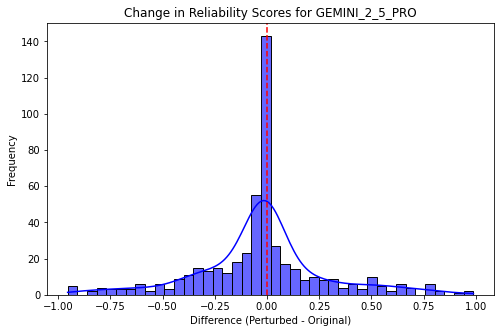}
\caption{Predicted reliability score differences in the results of GEMINI 2.5 PRO judge with two different prompts}
\label{fig:prompt_prediction_prob_diff}
\end{figure}

\pagebreak
The XGBoost judge reliability models are then trained using the scores from the second prompt. We observe changes in the
reliability scores across judges with the perturbed prompt. The distribution of the reliability score differences is shown
in Fig. \ref{fig:prompt_prediction_prob_diff}. It is seen that on average roughly $50\%$ of the datapoints have less than 0.1 difference in the reliability
scores of the judges.

\pagebreak
It is also observed that even though the scores might change by one score point, their agreement or disagreement with the
human given scores generally remains consistent even with slightly different prompts. For example, in the case of
Gemini 2.5 PRO as the judge, for 439/495 datapoints, the judge responses either both match the human score or both disagree.
Table \ref{tab:agreement_changes_due_prompt} summarizes the agreement changes with the human scores.

\begin{table}[h!]
\caption{Agreement changes with the human scores for two prompts on GEMINI 2.5 PRO judge}
\label{tab:agreement_changes_due_prompt}
\centering
\scriptsize
\begin{tabular}{|l|p{2cm}|p{2cm}|}
\hline
\textbf{Measurement Metric} & \textbf{Both or Neither Match} & \textbf{Only one matches} \\
\hline

Number of datapoints            & 439   & 56    \\
$\% tage$ age of datapoints     & 88.7  & 11.3  \\
\hline
\end{tabular}
\end{table}

\begin{figure}[h]
\centering
\includegraphics[width=1.0\textwidth]{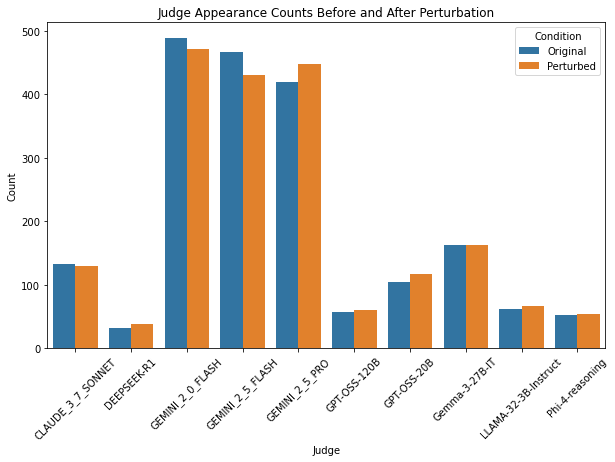}
\caption{Number of appearances of various judges in the selected jury panel}
\label{fig:judge_appearance_counts}
\end{figure}

We also observe how the number of appearances of the models in the selected jury panel changes
with slight change in the prompts. Fig. \ref{fig:judge_appearance_counts} shows that there is limited change in the inclusion of individual
judges in a given jury with the changed prompt.

\pagebreak
We observe that on average, 3.5 out of 4 final judges in the jury remained the same on using different prompts.

\begin{table}[h!]
\caption{Distribution of the number of common judges between the two runs with different prompts}
\label{tab:number_of_common_judges}
\centering
\scriptsize
\begin{tabular}{|l|p{1cm}|p{1cm}|p{1cm}|p{1cm}|p{1cm}|}
\hline
\textbf{$\#$ of common judges} & \textbf{0} & \textbf{1} & \textbf{2} & \textbf{3} & \textbf{4}\\
\hline

Counts     & 0 & 3 & 21 & 182 & 289\\
\hline
\end{tabular}
\end{table}

Table \ref{tab:number_of_common_judges} shows that the jury compositions do not change much when using slightly different prompts.

The perturbation analysis demonstrates that the evaluation framework is highly robust to prompt variations. While minor prompt
changes introduce some variability (about $30\%$ of data points shift by one score point), the overall agreement with human
evaluations remains largely stable. For instance, in the Gemini 2.5 PRO case, over $88\%$ of instances (439/495) either
consistently match or consistently diverge from human scores across prompt variations.

Reliability score changes exhibit minimal changes, with $50\%$ of data points showing less than 0.1 difference,
indicating that prompt-induced randomness is minimal and does not significantly affect judge reliability. Furthermore, jury
composition stability is striking. On average, 3.5 out of 4 jury members remain unchanged, and the aggregated jury scores
show negligible deviation. This suggests that the methodology maintains integrity even under prompt perturbations.

The system’s resilience to prompt changes implies that the evaluation pipeline is not overly sensitive to linguistic nuances,
making it suitable for real-world deployment where prompt variability is inevitable. This robustness ensures consistent
decision-making and fairness in model evaluation, reinforcing confidence in the methodology for broader applications.

\subsection{Model Weakness Analysis}\label{appendix:model_weakness}

To identify conditions where the jury model underperforms, we analyzed feature distributions between instances with high and low prediction discrepancies - excluding embedding features for interpretability. We focus on the RAG groundedness task for illustration. The dataset was split into two groups:\begin{enumerate} 
\item High Difference: Instances with large prediction errors. 
\item Low/No Difference: Instances with minimal errors. 
\end{enumerate}
For each feature, we computed mean and median values across both groups and ranked features by their median absolute differences in Table \ref{tab:model_weakness}. This revealed two failure modes:\begin{enumerate}
\item \textbf{Systematic Bias:} Features like factual density and named entities show consistent shifts in both mean and median, suggesting bias toward certain content structures. 
\item \textbf{Distributional Fragility:} Features such as syntactic anomaly and subjectivity show high median shifts but low mean differences, indicating sensitivity to rare or irregular linguistic patterns. 
\end{enumerate}

\begin{table}[h!]
\caption{Top 5 features ranked by median differences between high-difference and low/no-difference groups.}
\label{tab:model_weakness}
\centering
\begin{tabular}{|l|c|c|}
\hline
\textbf{Feature} & \textbf{Median Difference} & \textbf{Mean Difference} \\
\hline
utterance\_SYNTACTIC\_ANOMALY & 1.97 & 0.11 \\
utterance\_NAMED\_ENTITY & 0.77 & 0.37 \\
utterance\_SYNTACTIC\_AMBIGUITY & 0.74 & 0.22 \\
context\_FACTUAL\_DENSISTY & 0.63 & 0.36 \\
context\_SUBJECTIVITY & 0.53 & 0.17 \\
\hline
\end{tabular}
\end{table}

\section{Judge Scoring Runtimes}
\label{a:timing}

Here we provide some notes the runtime for judges scoring select data.   For this work, each judge was implemented in one of three distinct environments: Google Cloud API, Nvidia H200 with 140 GB RAM, or Nvidia TESLA V100 with 32 GB RAM. 
These choices were made based on the nature of the models (closed vs. open) and resource needs of the the open-weight models.

\begin{itemize}
\item GCP was selected for the closed models such as Claude 3.7 Sonnet and Gemini 2.5.
\item The Nvidia H200 was used for larger open-weight models like Phi-4 Reasoning, Llama 3.2-3B, and Gemma 3 27B-IT, leveraging its high memory capacity for large-scale inference.
\item V100 served as a baseline GPU environment for evaluating performance under constrained resources.
\end{itemize}

The runtime for each dataset varied depending on several factors including the size of the dataset, the computational environment used, and the specific model being executed.  Table \ref{tab:runtimes} summarizes the runtimes observed for select datasets during our experiments.  Note that this is not intended as to represent a rigorus comparitive study of the models but to provide a general sense of the time necessary for executing judge scoring.

\begin{table}[h!]
\centering
\begin{tabular}{|l|c|c|c|c|c|}
\hline
 \textbf{LLM} & \textbf{Environment} &\textbf{TLDR} & \textbf{unisumeval} & \textbf{MS MARCO} & \textbf{QASPER} \\
\hline
Claude 3.7 SONNET &  \multirow{{3}}{*}{GCP} &  3 mins & 4 mins & 7 mins & 5 mins \\
Gemini 2.5 Pro &     &  6 mins & 6 mins & 26 mins & 2 mins \\
Gemini 2.5 Flash &     &  2 mins & 3 mins & 13 mins & 2 mins \\
\hline
Phi-4 Reasoning &  \multirow{3}{*}{H200} &  81 mins & 116 mins & 579 mins & 36 mins \\
Llama 3.2-3B &      & 9 mins & 68 mins & 136 mins & 17 mins \\
Gemma 3 27B-IT &      & 17 mins & 66 mins & 477 mins & 17 mins \\
\hline
DeepSeek-R1 &   \multirow{1}{*}{V100} & 76 mins & 401 mins & 721 mins & 18 mins \\
\hline
\end{tabular}
\caption{Observed model runtimes (in minutes) across different environments and datasets used in the analysis}
\label{tab:runtimes}
\end{table}

\section{Framework Generalizability to Unseen Domains} \label{appendix:unseen_domain}

To assess the framework’s generalizability to unseen domains, we employ a leave-one-out procedure. Specifically, for each experiment, one data source is excluded from the training of XGBoost reliability models and jury construction, and the trained framework is then evaluated on the held-out source. This approach tests whether the Jury-on-Demand mechanism consistently outperforms both static juries and individual judges in previously unseen domains.

Using the RAG-Relevance task as an illustrative example, the dataset comprises three sources: ALCE, Hotpot-QA, and MS MARCO. In the first iteration, ALCE is held out while the framework is trained on Hotpot-QA and MS MARCO; performance is then assessed on ALCE. The process is repeated for each remaining source. The complete results are presented in Table \ref{tab:domain_RAG}. Across all three cases, Jury-on-Demand achieves the highest performance, indicating that the learned patterns generalize effectively to held-out domains.

\begin{table}[h!]
\caption{Kendall's tau on held-out data source - RAG Relevance}
\label{tab:domain_RAG}
\centering
\begin{tabular}{|p{0.8cm}p{0.8cm}p{0.8cm}p{0.9cm}p{0.8cm}p{0.8cm}p{0.8cm}p{0.6cm}p{0.6cm}p{0.5cm}p{0.8cm}p{0.5cm}p{0.7cm}|}
\hline
 Held-out & Jury-on-Demand & Static-Jury & Claude 3.7 & Gemn. 2.0 & Gemn. 2.5 & Gemn. 2.5 Pro & GPT-OSS-20B & GPT-OSS-120B & LL-3.2 & Gemma 3 & Phi-4 &  Deep-Seek-R1 \\
\hline
ALCE & 0.62 & 0.59 & 0.58 & 0.55 & 0.57 & 0.57 & 0.58 & 0.6 & 0.22 & 0.42 & 0.13 & 0.28 \\
Hotpot-QA  & 0.92 & 0.86 & 0.87 & 0.81 & 0.88 & 0.88 & 0.89 & 0.89 & 0.39 & 0.78 & 0.47 & 0.57  \\
MS-MARCO  & 0.46 & 0.38 & 0.44 & 0.38 & 0.4 & 0.43 & 0.39 & 0.4 & 0.2 & 0.12 & 0.12 & 0.29 \\
\hline
\end{tabular}
\end{table}

We extend our evaluation to the Summarization-Groundedness task, which includes four data sources: Summ-Eval, TL;DR, UniSum-Eval, and Dial-Summ-Eval. Table \ref{tab:domain_summ} reports the performance of the Jury-on-Demand framework under the leave-one-out setting for each source. The results indicate strong generalization for three sources, while performance on Summ-Eval is weaker.

These findings reinforce that the framework’s generalizability is influenced by both the diversity of training data and the characteristics of unseen domains. As additional annotated datasets become available and incorporated into training, we expect the framework’s ability to generalize to new domains to improve substantially.

\begin{table}[h!]
\caption{Kendall's tau on held-out data source - Summarization Groundedness}
\label{tab:domain_summ}
\centering
\begin{tabular}{|p{0.8cm}p{0.8cm}p{0.8cm}p{0.9cm}p{0.8cm}p{0.8cm}p{0.8cm}p{0.6cm}p{0.6cm}p{0.5cm}p{0.8cm}p{0.5cm}p{0.7cm}|}
\hline
 Held-out & Jury-on-Demand & Static-Jury & Claude 3.7 & Gemn. 2.0 & Gemn. 2.5 & Gemn. 2.5 Pro & GPT-OSS-20B & GPT-OSS-120B & LL-3.2 & Gemma 3 & Phi-4 &  Deep-Seek-R1 \\
\hline
Dial-Summ-Eval & 0.7 & 0.64 & 0.63 & 0.63 & 0.69 & 0.65 & 0.66 & 0.68 & 0.3 & 0.69 & 0.26 & 0.28 \\
\hline
Summ-Eval & 0.63 & 0.68 & 0.63 & 0.54 & 0.68 & 0.66 & 0.64 & 0.68 & 0.33 & 0.63 & 0.23 & 0.02 \\
\hline
TL;DR & 0.46 & 0.43 & 0.38 & 0.43 & 0.32 & 0.33 & 0.27 & 0.41 & 0.06 & 0.42 & 0.12 & 0.14 \\
\hline
Uni-Summ-Eval & 0.63 & 0.53 & 0.52 & 0.56 & 0.59 & 0.58 & 0.6 & 0.59 & 0.15 & 0.58 & 0.17 & -0.1 \\
\hline
\end{tabular}
\end{table}

\section{Judge Score Calibration} \label{appendix:judge_calibration}

\begin{figure}[htbp]
\centering
\begin{minipage}[b]{0.48\textwidth}
  \centering
  \includegraphics[width=\textwidth]{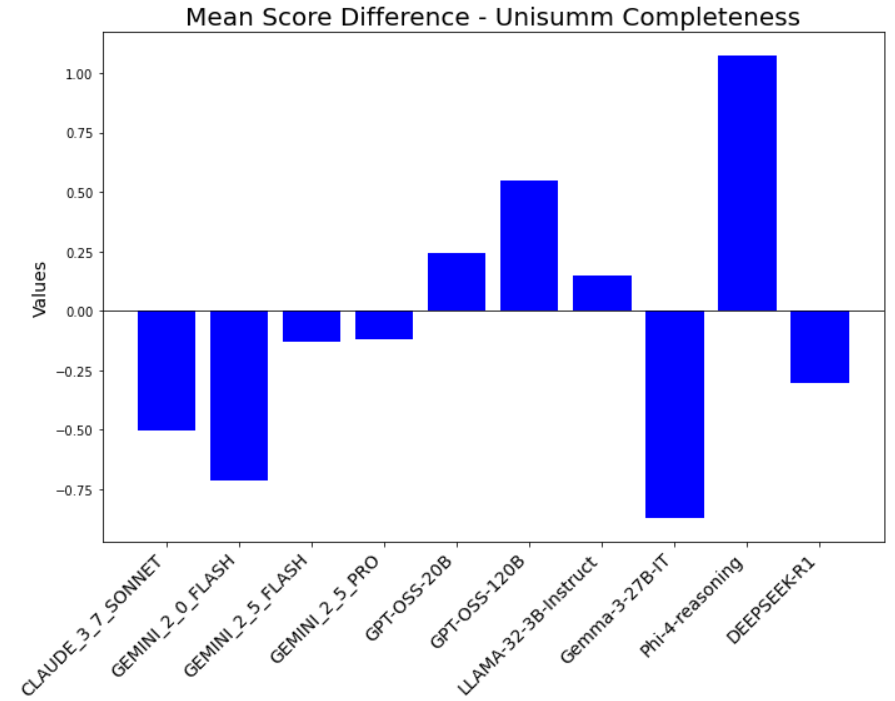}
\end{minipage}
\hfill
\begin{minipage}[b]{0.48\textwidth}
  \centering
  \includegraphics[width=\textwidth]{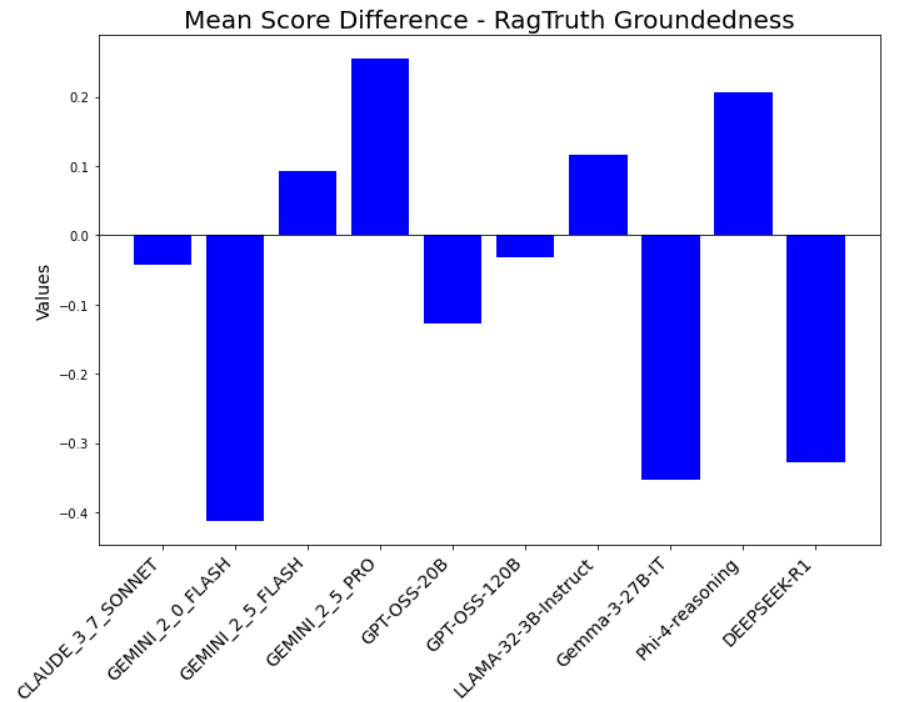}
\end{minipage}
\caption{The difference between each judge’s mean score and the mean human annotation score (annotation score minus judge score) for completeness and groundedness in Unisumm.}
\label{fig:calib_diff}
\end{figure}

Some judges may consistently assign lower or higher scores compared to human annotations. To address this, we apply score calibration. Specifically, we perform isotonic calibration \citep{niculescu2005predicting} for each judge's score within each dataset, then retrain the XGBoost model and construct the jury using the calibrated scores. For illustration, we focus on summarization completeness and RAG groundedness. Prior to calibration, we examine the difference between each judge’s mean score and the mean human annotation score (annotation score minus judge score). Fig. \ref{fig:calib_diff} show these differences for completeness in Unisumm and groundedness in RagTruth. We observe that weaker models—particularly LLAMA 3.2 3B Instruct, Gemma 3.2 7B IT, Phi 4 Reasoning and DeepSeek R1—tend to exhibit larger discrepancies. 

\begin{figure}[htbp]
\centering
\begin{minipage}[b]{0.48\textwidth}
  \centering
  \includegraphics[width=\textwidth]{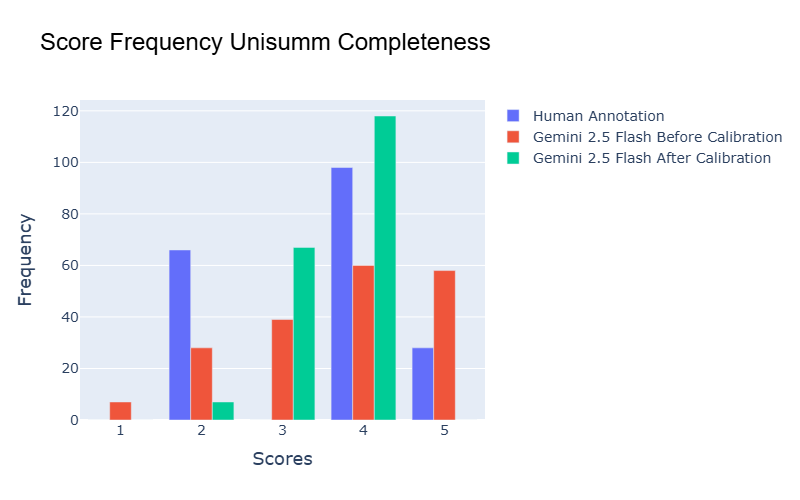}
\end{minipage}
\hfill
\begin{minipage}[b]{0.48\textwidth}
  \centering
  \includegraphics[width=\textwidth]{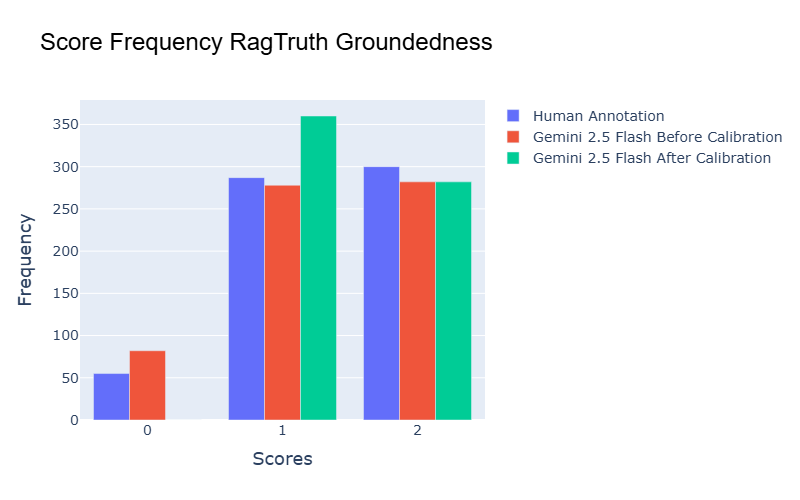}
\end{minipage}
\caption{The difference between each judge’s mean score and the mean human annotation score (annotation score minus judge score) for completeness and groundedness in Unisumm.}
\label{fig:calib_score}
\end{figure}

Tables \ref{tab:calib_cmpl} and \ref{tab:calib_grd} compare Kendall’s tau before and after calibration. Overall performance changes are minimal, with calibrated results slightly improving on RAG groundedness (0.69 vs. 0.67) but slightly worsening on summarization completeness (0.46 vs. 0.49). Differences become more pronounced within certain datasets and for specific judges. For example, as shown in Fig. \ref{fig:calib_score} (left), tau for Gemini 2.5 Flash on Unisumm completeness drops from 0.62 to 0.54 because many original score 5s are calibrated to 4, leading to underestimation of human annotation scores. Conversely, for RagTruth groundedness (Fig. \ref{fig:calib_score} (right)), Gemini 2.5 Flash’s tau increases from 0.57 to 0.58 after calibration because scores of 0 are calibrated to 1, reducing bias in judge scores. In summary, calibration can be beneficial for certain tasks and judges, but it may also introduce under- or overestimation of human annotations, reducing alignment. Future work will explore strategies to mitigate judge score bias more effectively.

\begin{table}[h!]
\caption{Summarization completeness: Kendall’s tau after calibration, with original values shown in parentheses. NA indicates cases where all scores were calibrated to a single value, making Kendall’s tau computation infeasible.}
\label{tab:calib_grd}
\centering
\begin{tabular}{|p{0.8cm}p{0.7cm}p{0.7cm}p{0.7cm}p{0.7cm}p{0.7cm}p{0.7cm}p{0.7cm}p{0.7cm}p{0.7cm}p{0.7cm}p{0.7cm}|}
\hline
Model & Jury-on-Demand & Claude 3.7 & Gemn. 2.0 & Gemn. 2.5 & Gemn. 2.5 Pro & GPT-OSS-20B & GPT-OSS-120B & LL-3.2 & Gma. 3 & Phi-4 & DS-R1\\
\hline
Overall & 0.46 & 0.38 & 0.47 & 0.51 & 0.44 & 0.38 & 0.46 & 0.15 & 0.22 & 0.16 & 0.23 \\
           & (0.49) & (0.39) & (0.46) & (0.46) & (0.43) & (0.39) & (0.47) & (0.13) & (0.25) & (0.22) & (0.27) \\\hline
Summ-Eval & 0.63  & 0.64 & 0.61 & 0.58 & 0.47 & NA & 0.62 & -0.04 & 0.07 & 0.31 & 0.69 \\
          & (0.72)  & (0.59) & (0.62) & (0.60) & (0.46) & (-0.02) & (0.63) & (-0.04) & (0.24) & (0.21) & (0.69) \\\hline
TL;DR & 0.35 & 0.31 & 0.41 & 0.36 & 0.35 & 0.29 & 0.39 & 0.06 & 0.23 & NA & 0.12 \\
          & (0.43) & (0.32) & (0.40) & (0.37) & (0.37) & (0.34) & (0.39) & (0.12) & (0.17) & (0.02) & (0.12) \\\hline
Uni-Summ-Eval & 0.58 & 0.41 & 0.46 & 0.54 & 0.46 & 0.66 & 0.61 & 0.26 & 0.54 & 0.33 & 0.35 \\
          & (0.61) & (0.57) & (0.58) & (0.62) & (0.58) & (0.62) & (0.71) & (0.26) & (0.64) & (0.34) & (0.38) \\\hline
\end{tabular}
\end{table}

\begin{table}[h!]
\caption{RAG groundedness: Kendall’s tau after calibration, with original values shown in parentheses. NA indicates cases where all scores were calibrated to a single value, making Kendall’s tau computation infeasible.}
\label{tab:calib_cmpl}
\centering
\begin{tabular}{|p{0.8cm}p{0.7cm}p{0.7cm}p{0.7cm}p{0.7cm}p{0.7cm}p{0.7cm}p{0.7cm}p{0.7cm}p{0.7cm}p{0.7cm}p{0.7cm}|}
\hline
Model & Jury-on-Demand & Claude 3.7 & Gemn. 2.0 & Gemn. 2.5 & Gemn. 2.5 Pro & GPT-OSS-20B & GPT-OSS-120B & LL-3.2 & Gma. 3 & Phi-4 & DS-R1\\
\hline
Overall & 0.69 & 0.56 & 0.50 & 0.64 & 0.60 & 0.63 & 0.63 & NA & 0.10 & 0.03 & 0.28 \\
           & (0.67) & (0.53) & (0.48) & (0.61) & (0.56) & (0.61) & (0.61) & (0.09) & (0.02) & (0.13) & (0.25) \\\hline
CAQA & 0.65  & 0.57 & 0.57 & 0.58 & 0.57 & 0.57 & 0.59 & NA & NA & NA & 0.19 \\
          & (0.65)  & (0.57) & (0.58) & (0.57) & (0.57) & (0.57) & (0.58) & (0.1) & (0.01) & (0.03) & (0.23)\\\hline
Halu-Eval & 0.79  & 0.72 & 0.52 & 0.79 & 0.76 & 0.78 & 0.79 & NA & 0.14 & -0.04 & 0.39 \\
          & (0.79)  & (0.70) & (0.51) & (0.77) & (0.76) & (0.78) & (0.79) & (0.14) & (0.14) & (0.11)  & (0.39)\\\hline
Rag-Truth & 0.55 & 0.48 & NA & 0.58 & 0.50 & 0.55 & 0.53 & NA & NA & NA & NA \\
          & (0.57) & (0.44) & (0.32) & (0.57) & (0.47) & (0.52) & (0.51) & (0.13) & (0.16) & (0.21) & (0.24) \\\hline
\end{tabular}
\end{table}

\section{Comparison with Baselines} \label{appendix:compare_baseline}

In this section, we examine how performance varies with the number of judges $K$, we focused on two tasks for illustration: Summarization-Completeness and RAG-Groundedness. We compared Jury-on-Demand (JOD) with two static jury baselines: Simple Average and Weighted Regression (see section \ref{sec:eval_protocol} for definitions). For these static jury baselines, we first determine which judges to include. Specifically, we rank judges by their Kendall’s tau performance for each task and select the top $K$ judges for the static jury. Table \ref{tab:judge_ranking} shows the judge ranked by performance for each task. Tables \ref{tab:sens_k_summ_cmpl} and \ref{tab:sens_k_summ_cmpl} illustrate how Kendall’s tau changes with jury size $K$ for Summarization-Completeness and RAG-Groundedness, respectively. Results are averaged over 10 runs. Our main observations are:

\begin{itemize}
\item JOD consistently outperforms static juries: Performance varies with jury size $K$, but JOD consistently outperforms both static baselines across all $K$ values, except for a few cases in Summarization-Completeness. The performance margin of JOD over static juries is larger for RAG-Groundedness than for Summarization-Completeness.

\item Effect of jury size differs by task: For Summarization-Completeness, there is no clear pattern on performance do not for different juries. JOD's performance do not vary too much as $K$ changes, Simple Average jury show performance decrease as $K$ increases. For RAG-Groundedness, performance generally improves with larger $K$, except that Simple Average shows a drop at $K = 10$. These observations can be exaplained by the larger score bias in weaker judges. Large biases degrade performance when these judges are included, especially for Simple Average, which assigns equal weight regardless of bias.

\item Weighted Regression struggles with small K: Weighted Regression performs worse than other jury methods when $K$ is small, particularly for RAG-Groundedness. One contributing factor is that the regression coefficients for RAG-Groundedness are very small, which results in weighted regression scores that fall below human annotation scores. To ensure realistic outputs, we round up any weighted regression score that is lower than the minimum human annotation score. However, because the coefficients are small, this adjustment often causes many weighted regression scores to equal the lowest human annotation score, further degrading performance. Another potential reason for poor performance at small $K$ is the high correlation among strong judges (e.g., Gemini models, GPT-OSS models, Claude 3.7 Sonnet), with correlation values around 0.7–0.8. Using such highly correlated features in regression can negatively impact model performance.
\end{itemize}

In summary, how performance changes with jury size $K$ depends on multiple factors, including task characteristics, score distributions, and correlations among judge scores. Although JOD requires more training data and a more complex training process compared to static juries, its dynamic jury selection and adaptive weight assignment enable it to choose the best judges for each instance and achieve superior overall performance.

\begin{table}[htbp]
\centering
\scriptsize
\caption{Judges ranked by Kendall's tau (high to low).}
\label{tab:judge_ranking}
\begin{tabular}{p{1.3cm}|p{0.8cm}p{0.8cm}p{0.8cm}p{0.8cm}p{0.8cm}p{0.8cm}p{0.8cm}p{0.8cm}p{0.8cm}p{0.8cm}}
\toprule
\textbf{K} & \textbf{1} & \textbf{2} & \textbf{3} & \textbf{4} & \textbf{5} & \textbf{6} & \textbf{7} & \textbf{8} & \textbf{9}  & \textbf{10}\\
\midrule
Summ-complete-ness & GPT-OSS-120B & Gemn-2.5 & Gemn-2.0 & Gemn-2.5-Pro & Claude & GPT-OSS-20B & DS-R1 & Gemma-3 & Phi 4 & LL 3.2\\ \hline
RAG-grounded-ness           & GPT-OSS-120B & Gemn-2.5 & GPT-OSS-20B & Gemn-2.5-Pro & Claude & Gemn-2.0 & DS-R1 & Phi 4 & LL 3.2 & Gemma-3\\
\bottomrule
\end{tabular}
\end{table}

\begin{table}[htbp]
\centering
\scriptsize
\caption{Summarization Completeness: Kendall’s tau across varying jury sizes for different baselines.}
\label{tab:sens_k_summ_cmpl}
\begin{tabular}{p{1.8cm}|p{0.9cm}p{0.8cm}p{0.8cm}p{0.8cm}p{0.8cm}p{0.8cm}p{1.2cm}p{0.8cm}p{1cm}p{0.8cm}}
\toprule
\textbf{K}  & \textbf{2} & \textbf{3} & \textbf{4} & \textbf{5} & \textbf{6} & \textbf{7} & \textbf{8} & \textbf{9}  & \textbf{10}\\
\midrule
Jury-on-Demand  & 0.43 & 0.45 & 0.45 & 0.46 & 0.46 & 0.47 & 0.47 & 0.46 & 0.46\\ \hline
Static (Average K)  & 0.46 & 0.47 & 0.48 & 0.48 & 0.46 & 0.45 & 0.45 & 0.44 & 0.43\\ \hline
Static (Average K)  & 0.46 & 0.45 & 0.45 & 0.46 & 0.46 & 0.46 & 0.46 & 0.44 & 0.45\\
\bottomrule
\end{tabular}
\end{table}

\begin{table}[htbp]
\centering
\scriptsize
\caption{RAG Groundedness: Kendall’s tau across varying jury sizes for different baselines.}
\label{tab:sens_k_rag_grd}
\begin{tabular}{p{1.8cm}|p{0.9cm}p{0.8cm}p{0.8cm}p{0.8cm}p{0.8cm}p{0.8cm}p{1.2cm}p{0.8cm}p{1cm}p{0.8cm}}
\toprule
\textbf{K}  & \textbf{2} & \textbf{3} & \textbf{4} & \textbf{5} & \textbf{6} & \textbf{7} & \textbf{8} & \textbf{9}  & \textbf{10}\\
\midrule
Jury-on-Demand  & 0.65 & 0.67 & 0.67 & 0.68 & 0.68 & 0.67 & 0.68 & 0.67 & 0.66\\ \hline
Static (Average K)  & 0.64 & 0.65 & 0.65 & 0.65 & 0.65 & 0.64 & 0.64 & 0.63 & 0.58\\ \hline
Static (Average K)  & 0.58 & 0.58 & 0.56 & 0.54 & 0.52 & 0.60 & 0.65 & 0.65 & 0.65\\
\bottomrule
\end{tabular}
\end{table}

\section{Parameter Tuning in XGBoost} \label{appendix:parameter_tuning}

We use random search to tune the hyperparameters of the XGBoost models for judge reliability scores. The search space for the tuned parameters is provided in Table \ref{tab:parameter_xgboost}. Parameters not listed in the table are set to their default values.

\begin{table}[h!]
\caption{Parameter search space in XGBoost.}
\label{tab:parameter_xgboost}
\centering
\begin{tabular}{|l|c|}
\hline
\textbf{Parameter} & \textbf{Search Space} \\
\hline
max\_depth      & 2, 3, 4, 5, 6, 7, 8, 9 \\
learning\_rate  & 0.01, 0.03, 0.05, 0.07, 0.1, 0.2 \\
n\_estimators   & 500, 600, 800, 1000, 1200 \\
\hline
\end{tabular}
\end{table}

\end{document}